\documentclass[conference]{IEEEtran}
\usepackage[font=small,format=plain,labelfont=bf,textfont=bf]{caption}
\usepackage[export]{adjustbox}
\usepackage[colorlinks=true, allcolors=blue]{hyperref}
\usepackage{cite}
\usepackage[utf8]{inputenc}
\usepackage{xcolor}
\usepackage[T1]{fontenc}
\usepackage{url}
\usepackage{nicefrac}
\usepackage{graphicx}
\usepackage{multirow}
\usepackage{amsmath, amssymb, amsfonts}
\usepackage{amsthm}
\usepackage{mathrsfs}
\usepackage{textcomp}
\usepackage{manyfoot}
\usepackage{booktabs}
\usepackage{algorithm}
\usepackage{algorithmicx}
\usepackage{algpseudocode}
\usepackage{listings}
\usepackage{enumitem}
\usepackage{subcaption}
\usepackage{tabularx}
\usepackage[most]{tcolorbox}
\usepackage[capitalise]{cleveref}
\usepackage{newtxmath}
\usepackage{bm}
\usepackage{indentfirst}
\usepackage[framemethod=TikZ]{mdframed}

\usepackage{caption}
\usepackage[font=small,format=plain,labelfont=bf,textfont=bf]{caption}
\usepackage[export]{adjustbox}
\usepackage[colorlinks=true, allcolors=blue]{hyperref}
\usepackage{cite}
\usepackage[utf8]{inputenc}
\usepackage{xcolor}
\usepackage[T1]{fontenc}
\usepackage{url}
\usepackage{nicefrac}
\usepackage{graphicx}
\usepackage{multirow}
\usepackage{amsmath, amssymb, amsfonts}
\usepackage{amsthm}
\usepackage{mathrsfs}
\usepackage{textcomp}
\usepackage{manyfoot}
\usepackage{booktabs}
\usepackage{algorithm}
\usepackage{algorithmicx}
\usepackage{algpseudocode}
\usepackage{listings}
\usepackage{enumitem}
\usepackage{subcaption}
\usepackage{tabularx}
\usepackage[most]{tcolorbox}
\usepackage[capitalise]{cleveref}
\usepackage{newtxmath}
\usepackage{bm}
\usepackage{indentfirst}
\usepackage{todonotes}
\usepackage[framemethod=TikZ]{mdframed}

\newtcolorbox{findingbox}{
  colback=gray!20,
  colframe=white,
  left=2mm,
  right=2mm,
  top=0.7mm,
  bottom=0.7mm,
  boxsep=0mm,
  arc=0mm
}

\newcommand{\smartparagraph}[1]{\noindent{\bf #1}\ }

\crefname{section}{\S}{\S\S}

\DeclareMathOperator{\Dir}{Dir}

\title{Analysis and Tuning of Knowledge Distillation for Efficient Collaborative Learning}

\author{\IEEEauthorblockN{Norah Alballa}
\IEEEauthorblockA{
\textit{KAUST}\\
}
\and
\IEEEauthorblockN{Ahmed M. Abdelmoniem}
\IEEEauthorblockA{
\textit{Queen Mary University of London}\\
}
\and
\IEEEauthorblockN{Marco Canini}
\IEEEauthorblockA{
\textit{KAUST}\\
}
}

\begin{document}

\IEEEtitleabstractindextext{

\begin{abstract}

Collaborative learning addresses key challenges in distributed systems, such as large-scale training, data privacy, and computational heterogeneity, by enabling decentralized entities to collaborate without centralizing data. This distributed computing paradigm is particularly crucial for heterogeneous environments, where maintaining communication efficiency and supporting diverse model architectures are vital. Among the various approaches to collaborative learning, knowledge distillation (KD) has emerged as a promising method, facilitating efficient knowledge transfer while minimizing communication overhead. However, existing research on KD often lacks a systematic empirical analysis of its effectiveness across distributed and heterogeneous settings.

This paper bridges this gap by conducting a comprehensive experimental analysis of KD methods for distributed systems. We evaluate vanilla KD, tuned KD, deep mutual learning, and data-partitioning KD across diverse data distributions and system configurations. Our results demonstrate that tuning KD hyperparameters, such as temperature and weight, enhances accuracy gains by 16\% in uniform distributions and achieves an average improvement of 139\% across non-IID scenarios compared to the best non-tuned approach. 
Furthermore, we propose a novel solution
which leverages KD to enhance training efficiency in centralized collaborative learning and FL settings. By consolidating teacher models into a unified starting point, this approach achieves an average reduction of 60.6\% in total communication rounds, effectively accelerating the training process. 
These findings provide insights and practical guidelines for collaborative learning in distributed systems.

\end{abstract}

}

\maketitle

\IEEEdisplaynontitleabstractindextext

\IEEEpeerreviewmaketitle

\section{Introduction}\label{sec:intro}
\IEEEPARstart{K}{nowledge} Distillation, initially conceptualized for model compression \cite{hinton2015distilling,bucilua2006model,schmidhuber1991neural}, 
has evolved to facilitate knowledge transfer between pre-trained models\cite{li2021decentralized,li2019fedmd,itahara2021distillation,sattler2021cfd}. 
Notably, these pre-trained models might be exposed to distinct (i.e., unseen or out of distribution) data samples, resulting in performance variations due to various types of heterogeneity. One of these is statistical heterogeneity, where models are trained with data that are not independent and identically distributed (non-IID). Additionally, system heterogeneity introduces challenges as devices vary in their available bandwidth and computational capabilities. Learning across such diverse systems and imbalanced datasets is complex yet critical.

The application of KD in decentralized collaborative learning and federated learning environments is increasingly recognized for its potential to mitigate communication overhead and facilitate diverse model architectures \cite{itahara2021distillation, sattler2021cfd, sattler2021fedaux, li2021decentralized, li2019fedmd, kalra2021proxyfl, hu2021mhat, chang2019cronus, REFL_2023, Amna_ComCom_2024}. Despite this growing interest, a notable gap exists in understanding KD's application, particularly in pre-trained models. Recent studies employ various KD algorithms—such as vanilla KD, deep mutual learning \cite{li2021decentralized,kalra2021proxyfl}, and data partitioning KD \cite {roth2023fantastic}—to facilitate knowledge transfer between pre-trained models.  
These methods' effectiveness across varied data distribution strategies and their comparative advantages or limitations have not been systematically and empirically investigated.  Additionally, KD hyperparameters tuning, particularly concerning parameters such as temperature and weight, remains largely unexplored, presenting a significant opportunity for advancement in this domain \cite{alballa2023first}.

This paper addresses these gaps by conducting an in-depth study on joint (or collaborative) distillation. In this scenario, distributed participants use their pre-trained models to engage in direct peer-to-peer knowledge exchange. We present a comparative analysis of diverse KD approaches, considering various data partitioning strategies. Our exploration includes an assessment of the influence of transfer set options and a detailed investigation into hyperparameter optimization. Through extensive grid search results, we identify practical guidelines for tuning to enhance model performance. This study aims to enrich the understanding of KD techniques in pre-trained models and provide actionable insights for optimizing these methods in distributed and collaborative learning contexts.

This paper makes the following contributions to the field of knowledge distillation for collaborative learning systems:

\begin{itemize}
  \item We conduct a comprehensive comparative analysis of KD approaches across five data partitioning strategies and assess the impact of different knowledge transfer set methods and hyperparameters settings.\footnote{We plan to release the source code associated with this study openly.}
  \item We examine hyperparameters optimization, focusing on temperature and weight, using grid search to ascertain when tuning is most beneficial and to determine optimal tuning scenarios for diverse data partitions.
  \item We explore KD learning dynamics, analyzing learning-forgetting balance and criteria for effective teacher model selection across data scenarios.
   \item We explore the criteria for effective teacher model selection across data scenarios.
  \item We demonstrate the efficiency of KD in the centralized collaborative learning and FL settings, evidenced by reduced training rounds to reach desired accuracy levels.
  \end{itemize}

\section{Background and Related Work}
\label{background}
This section introduces the concept of collaborative learning, elaborates on the foundations of KD, and discusses related work in these fields.
\vspace{-2mm}
\subsection{Collaborative Learning}

Collaborative learning is a paradigm where multiple models or agents work together to enhance collective performance through shared learning experiences. This model allows each device or node to learn independently while also enabling selective collaboration by transferring knowledge when necessary \cite{daga2019cartel}. Collaborative learning encompasses both centralized approaches, 
such as FL \cite{mcmahan2017communication}, and decentralized peer-to-peer approaches \cite{daga2019cartel,daga2023clue,lu2019collaborative,chang2019cronus,roth2023fantastic}.

Collaborative learning addresses critical challenges such as data privacy, bandwidth limitations, and data locality. However, it faces significant hurdles due to system and data heterogeneity \cite{ye2023heterogeneous,yao2022edge,Li_2020, NEURIPS2022_cf326db2,ghosh2019robustfederatedlearningheterogeneous,FLstudy_IOTJ_2023}. 
System heterogeneity arises from variations in hardware capabilities, network conditions, and available computational power across devices. This disparity can make it challenging to fit the same model across all participating devices uniformly and often leads to inefficient training processes, compromising the learning performance. Additionally, data heterogeneity—characterized by the non-IID nature of data distribution across different nodes—further complicates the scenario by producing models that do not generalize well across varied data distributions \cite{kairouz2021advances,mcmahan2017communication,REFL_2023}.

\vspace{-2mm}
\subsection{Knowledge Distillation}

KD is a technique initially proposed to transfer knowledge from a large, complex model (termed the \emph{teacher}) to a smaller model (termed the \emph{student}) to maintain comparable performance \cite{hinton2015distilling,bucilua2006model,schmidhuber1991neural}. The central concept is that the student model can learn more effectively by emulating the soft targets, which are the output probabilities computed using the softmax function parameterized by temperature \( T \), as shown:
\vspace{-1mm}
\begin{equation} 
\label{eqn:distillation}
   p_{i}^{t}(z_{t,l}, T) = \frac{\exp(z_{t,i,l} / T)}{\sum_j \exp(z_{t,j,l} / T)}.
\end{equation}

Here, $z_{t,i,l}$ and $z_{t,j,l}$ are logits for classes $i$ and $j$ from the teacher model for a specific data sample \( l \).  The temperature \( T \) is used to control the smoothness of the probability distribution; in KD contexts, it is often set to a value greater than 1 to produce softer probability distributions \cite{hinton2015distilling}.

\begin{figure}[t!]
\centering
\includegraphics[width=0.43\textwidth]{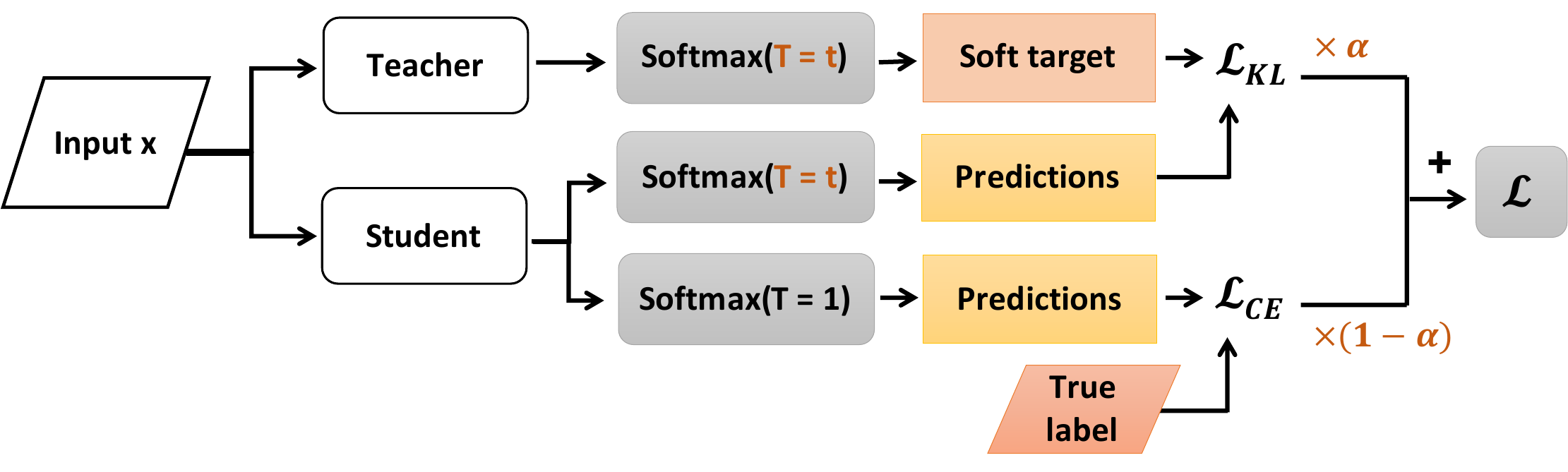}
\caption{\label{fig:kd} Knowledge distillation pipeline. 
}
\end{figure}

Various KD methodologies utilize different loss functions and datasets, often referred to as a \emph{transfer set} or \emph{proxy dataset}, which can include publicly labelled or unlabeled datasets or the teacher's training data. When labelled data is available, the student model's training combines the cross-entropy loss with the teacher's soft targets and a distillation loss as:
\begin{equation} 
\label{eqn:distillation-loss}
  \textstyle \mathcal{L} = (1 - \alpha) \: \mathcal{L}_{CE}(p^s,y) + \alpha \: \mathcal{L}_{KL}(p^{s}, p^{t}).
\end{equation}
The weight \( \alpha \) balances the cross-entropy loss \( \mathcal{L}_{CE} \) and the Kullback-Leibler divergence loss \( \mathcal{L}_{KL} \); \( y \) represents the true labels; \( p^{s} \) and \( p^{t} \) denote the predicted probabilities (cf. \cref{eqn:distillation}) by the student and teacher models, respectively. \cref{fig:kd} illustrates the KD pipeline.

\subsection{Knowledge Distillation With Pre-trained Models}
Beyond model compression, KD's applicability includes scenarios where both the teacher and student models are pre-trained on different data subsets. This application of KD aims to consolidate the distinct knowledge each model possesses, a concept increasingly relevant in decentralized collaborative learning ~\cite{chang2019cronus,sattler2021cfd, roth2023fantastic} and federated learning \cite{li2021decentralized,li2019fedmd,kalra2021proxyfl,itahara2021distillation,roth2023fantastic}.

Therefore, our study focuses on the analysis of the following knowledge distillation (KD) approaches:
\paragraph{Vanilla KD}
In the context of pre-trained models, Vanilla KD employs the standard KD process, (\cref{eqn:distillation-loss}), 
where the primary focus is on transferring knowledge from one pre-trained model to another, leveraging the established KD framework.

\paragraph{Deep Mutual Learning (DML)}
DML, also known as online or co-distillation, offers a perspective where two models, designated as student models \( s_1 \) and \( s_2 \), engage in a mutual learning process \cite{zhang2018deep,li2021decentralized,anil2018large}. Unlike the traditional teacher-student dynamic in KD, DML involves a peer-to-peer collaborative learning approach:
\vspace{-1.5mm}
\begin{align} 
\mathcal{L}_{s_1} = \mathcal{L}_{CE}(p^{s_1}, y) + \mathcal{L}_{KL}(p^{s_1}, p^{s_2}); \nonumber \\
\mathcal{L}_{s_2} = \mathcal{L}_{CE}(p^{s_2}, y) + \mathcal{L}_{KL}(p^{s_2}, p^{s_1}).
\end{align}

Here, losses $\mathcal{L}_{s_1}$ and $\mathcal{L}_{s_2}$ for student models $s_1$ and $s_2$ combine cross-entropy with true labels $y$ and KL divergence between each other's probabilities.

\paragraph{Data Partitioning KD (DP-KD)}

DP-KD \cite{roth2023fantastic} partitions the transfer data into two categories: one where the student model \( s \) benefits from the teacher model \( t \)'s feedback, and another where the student's pre-existing knowledge, represented by a frozen version of the initial student model (denoted as \( st \)), is preserved. Each data sample \( l \) is assigned to one of these categories based on the highest prediction probability for the corresponding ground-truth class:

\begin{equation} 
\begin{aligned}
m^{t}_{l} &= \textstyle\prod_{i} [p_{i}^{t}(z_{t,l}) > p_{i}^{st}(z_{st,l})]; \\
m^{st}_{l} &= \textstyle\prod_{i} [p_{i}^{t}(z_{t,l}) \leq p_{i}^{st}(z_{st,l})].
\end{aligned}
\end{equation}
In this context, \( m^{t}_{l} \) and \( m^{st}_{l} \) are binary masks for each sample \( l \), determining whether knowledge transfer from the teacher model \( t \) is appropriate, or if the student model \( s \) should retain its own knowledge. The distillation loss is defined as:

\begin{equation} 
\label{eqn:dp-kd}
\begin{aligned}
\textstyle \mathcal{L}_{dist} = & \frac{T^{2}}{n} \sum_{l=0}^{n} \Big[ m^{t}_{l} \mathcal{L}_{KL} \left( p^{s}(z_{s,l}, T), p^{t}(z_{t,l}, T) \right) \\
& + m^{st}_{l} \mathcal{L}_{KL} \left( p^{s}(z_{s,l}, T), p^{st}(z_{st,l}, T) \right) \Big].
\end{aligned}
\end{equation}

In the unsupervised variant of DP-KD, samples are assigned based on the maximum prediction probability for a sample, selecting the model with the highest confidence in its prediction. This model selection by confidence circumvents the need for true labels, enabling knowledge transfer even without labeled data. We explore the efficacy of both the supervised and unsupervised paradigms of DP-KD to assess their respective impacts on the performance of pre-trained models. 

\subsection{Related work}
\label{related}
\vspace{-1mm}

To address the limitations of system and data heterogeneity in collaborative learning and reduce communication overhead, KD has seen notable expansion, particularly in knowledge transfer among pre-trained models, as evidenced by a range of studies \cite{li2021decentralized, kalra2021proxyfl, hu2021mhat, sattler2021cfd}. Techniques such as FedMD \cite{li2019fedmd}, Cronus \cite{chang2019cronus}, and DS-FL \cite{itahara2021distillation} leverage KD to improve communication efficiency and address model heterogeneity challenges. Furthermore, decentralized learning frameworks like Def-KT \cite{li2021decentralized} and ProxyFL \cite{kalra2021proxyfl} employ online distillation to facilitate collaborative learning without a central authority.

Despite these advancements, much of the research remains focused on singular KD strategies and a narrow range of data partitioning methods. Our study seeks to fill this gap by examining a variety of KD methodologies across different data distribution and transfer set scenarios, aiming to detail the performance nuances of diverse KD techniques in distributed training. This includes exploring the impact of hyperparameter tuning in KD, as highlighted by recent findings \cite{alballa2023first}. Furthermore, while existing FL studies have successfully integrated KD to reduce communication loads \cite{itahara2021distillation, sattler2021cfd, sattler2021fedaux, li2021decentralized} and support flexible model architectures across participants \cite{li2019fedmd, kalra2021proxyfl, hu2021mhat, chang2019cronus}, our study proposes a new approach to further cut communication overhead, particularly in cross-silo FL settings, enhancing FL system efficiency.

\section{Methodology}
\label{sec:methodology}

\smartparagraph{Overview.}
Our methodology is structured to evaluate KD in collaborative learning environments. This research systematically compares different KD approaches under diverse scenarios, aiming to enhance our understanding of KD's application between pre-trained models, pinpoint the most effective KD strategies for specific conditions, and address existing gaps in the literature.
Our experimental framework begins with a foundational study in a controlled peer-to-peer setting involving a direct KD interaction between one student and one teacher. Subsequently, (in \ref{FL}), we extend the scope to include scenarios involving multiple teachers, simulating more complex, centralized collaborative learning settings and federated learning environments.

\subsection{Proposed Approach}

We aim to advance KD for collaborative learning by analyzing under-explored aspects and proposing novel solutions to address efficiency challenges in distributed settings. Our focus includes three key techniques: Tuned KD, Model Amalgamation, and Adaptive Teacher Weighting, each evaluated to provide actionable insights. Additionally, we introduce a novel pre-consolidation method for federated learning, which is discussed in later sections.

\smartparagraph{Tuned KD.} In this work, we study the tuning of KD hyper-parameters. Hence, our proposed approach, Tuned KD, refers to the optimization of KD hyperparameters, particularly the temperature (\( T \)) and weight (\( \alpha \)) hyper-parameters. This approach adjusts the standard KD framework to fit better the characteristics and learning capacity of the pre-trained student model. We aim to leverage this approach to systematically assess KD’s efficacy and adaptability across different collaborative configurations.

\smartparagraph{Model Amalgamation.}
We also study combining multiple models to amalgamate knowledge across multiple pre-trained models via KD effectively (which is analogous to ensemble learning). Our focus includes critical aspects such as selecting an appropriate starting student model and assessing various teacher weighting strategies across different transfer set scenarios.

\smartparagraph{Weighing Teachers.}
Traditionally, \emph{equal weighting} is used for all teachers; we investigate \emph{adaptive weighting}, which is based on each teacher's class-wise contribution. This weighting mechanism was inspired by a method outlined in the study \cite{aljahdali2024flashback}, adapted to use per-class accuracy instead of per-class label counts to ensure generalizability when label count information is unavailable.

In adaptive weighting, the weight vector \( \alpha_j \) for each teacher \( j \) considers their accuracy \(Acc_j^c\) for each class \( c \), relative to the student's accuracy \(Acc_s^c\) and the total accuracy of all teachers for that class:
\begin{equation}
\label{eqn:dw_multi}
\alpha_j^c = 
\begin{cases}
  \frac{Acc_j^c}{Acc_s^c + \sum_{i=1}^{J} Acc_i^c},& \text{if } Acc_s^c + \sum_{i=1}^{J} Acc_i^c > 0 \\
    0, & \text{otherwise}
\end{cases}
\end{equation}

Here, \(J\) is the total number of teachers, and the dimension of each \( \alpha_j \) vector matches the number of classes. The \( \mathcal{L}_{KL} \) in \cref{eqn:distillation-loss} can be updated considering the combined influence of multiple teachers, as:
\begin{equation}
\label{eqn:distillation-multi-kl}
   \textstyle \mathcal{L}_{KL_{multi}} = \sum_{j=1}^{J} \alpha_j \: \mathcal{L}_{KL}(p^{s}, p^{j}).
\end{equation}
This equation integrates the KL divergence losses between the student's predictions and each teacher's predictions, each scaled by its respective \( \alpha_j \). 

\begin{figure}[!t]
    \centering
    \includegraphics[width=0.38\textwidth]{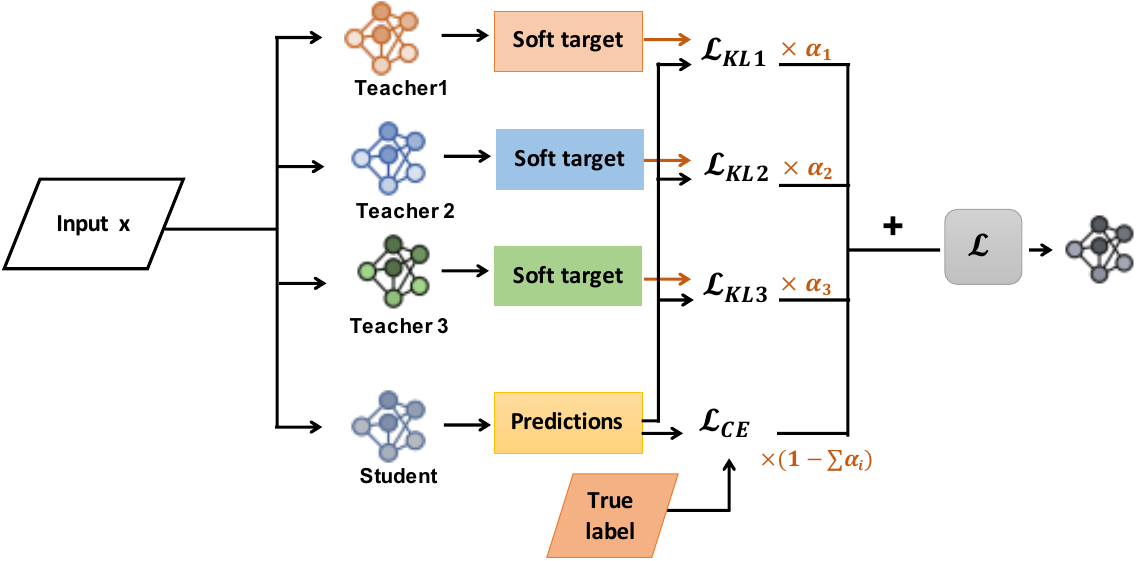}
    \caption{Pre-Consolidation of Models using KD. Multiple teacher models contribute their knowledge to create a consolidated model, which is used as the initial global model for the FL process.}
    \label{fig:pre_consolidation}
\end{figure}

\subsection{Evaluation Method}
\smartparagraph{Experimental Setup.}
Based on the assumption of a set of participants (so-called clients), each with a private dataset, this study focuses on the application of KD techniques between pre-trained models, performed as follows:
\begin{enumerate} [itemsep=0pt, parsep=0pt]
    \item Each model undergoes pre-training on its respective participant's training dataset. 
    \item Knowledge is then transferred from the teacher to the student model utilizing one of the aforementioned KD techniques.
    \item By default, the student model's training dataset is the transfer set. We also examine the use of public datasets, both labelled and unlabeled, to evaluate their impact on the distillation process.
\end{enumerate}

We assume a distributed training scenario with 10 participants for each data partitioning strategy. This setup results in 90 possible teacher-student pairings for each scenario, excluding self-transfers. 
We now describe the setup for this experimental methodology in detail.

\smartparagraph{Task, models, and datasets.}
This study focuses on a standard computer vision task, i.e., image classification.
As a dataset, we use CIFAR-10 \cite{cifar10} and CINIC-10 \cite{darlow2018cinic10}, which are commonly used in the FL literature (e.g., \cite{luo2021no,he2020group,duan2020self,lin2020ensemble,gong2021ensemble,bistritz2020distributed,lai2022fedscale,beutel2022flower}). 
The CIFAR-10 dataset consists of 32x32 color images of 10 classes. There are 50K training images and 10K test images. We use 10\% of the training set as a validation set. 
The CINIC-10 dataset comprises 270,000 images divided into training, validation, and test sets, each containing 90,000 images across 10 classes. We merge the training and validation sets, utilizing 1\% of this set for validation and the remaining 99\% for training. 
The testing set of each dataset is used to measure the accuracy of the participants' models before and after distillation. CIFAR-10 is the default data set, unless otherwise specified.

We utilize the ResNet-18 architecture \cite{he2016deep} for all participants to assess the effects of knowledge distillation techniques and data strategies employed, thereby eliminating confounding variables introduced by differing architectures.

We pre-train the model over its training dataset for at most 100 epochs for each participant. We employ early stopping after the validation performance plateaus for 10 epochs. 
We use the Adam optimizer with a learning rate of $1 \times 10^{-3}$, weight decay of $4 \times 10^{-4}$, and a batch size 32. These hyperparameters remain constant across all experiments.

\smartparagraph{Data partitioning strategies.}
We experiment with the following five data partitioning strategies:
\begin{enumerate}[itemsep=0pt, parsep=0pt]
   
    \item \textbf{Uniform Data Distribution:} Each participant \( P_i \) possesses an equal number of data samples from all classes, resulting in balanced local datasets \( D_i \), which promote uniform model training across the network.
    \item \textbf{Quantity Skew Distribution} \cite{li2022federated}: The size of each participant's local dataset \( |D_i| \) varies. A Dirichlet distribution, denoted as \( q_i \sim \Dir(\beta) \), is used to model the size variability, where \( q_i \) represents the proportion of data for participant \( P_i \). The concentration parameter \( \beta \), set to 0.5, controls the degree of data quantity skew among participants.
    \item \textbf{Specialized Non-IID Distribution:} In this strategy, each participant \( P_i \) primarily holds data from one class, making up 91\% of their dataset. The remaining 9\% is distributed evenly among samples from other classes, ensuring exposure to the entire class spectrum.
    \item \textbf{Label Skew Random Chunks Distribution:} This approach involves assigning a random number of classes to each participant, followed by allocating random chunks, each containing a minimum of 100 samples, from these classes. This leads to a controlled variation in class distribution and dataset volume.
    \item \textbf{Label Skew Dirichlet Distribution:} For each class \( c \), we construct a distribution vector \( p_c \sim \Dir_K(\boldsymbol{\beta}) \), where \( K \) is the number of participants and \( \boldsymbol{\beta} \) is a vector of concentration parameters alternating between 0.1 and 0.5. The proportions \( p_{c,k} \) derived from \( p_c \) determine the allocation of class \( c \)'s samples to participant \( P_k \), creating a skewed label distribution across participants, with lower \( \beta \) values leading to greater skew.
\end{enumerate}

\smartparagraph{KD setup.}
We assess the effectiveness of three distinct KD methodologies: \textit{Vanilla KD}, \textit{DML}, and \textit{DP-KD}. Each of these methodologies is explored with a trio of transfer set options:

\begin{enumerate} [itemsep=-0pt, parsep=0pt]
\item A public labeled dataset \cite{roth2023fantastic,cheng2021fedgems,zhang2021parameterized,hinton2015distilling}.
\item A public unlabeled dataset \cite{itahara2021distillation,chang2019cronus,sattler2021cfd,gong2021ensemble,sattler2021fedaux,chen2021fedbe,bistritz2020distributed,lin2020ensemble}.
\item The student's own training dataset \cite{alballa2023first,li2021decentralized}.
\end{enumerate}

For Vanilla KD, the parameters used are $T=1$ and $\alpha=0.5$, reflecting common practices in existing literature \cite{chang2019cronus,sattler2021cfd,gong2021ensemble,sattler2021fedaux,chen2021fedbe,li2021decentralized,kalra2021proxyfl}. When utilizing a public dataset as a transfer set, we incorporate a copy of the student's pre-trained model as an additional teacher, based on recommendations from \cite{alballa2023first}.

In the Tuned KD approach, we primarily focus on optimizing the $\alpha$ parameter, followed by tuning the temperature parameter $T$.  This optimization is conducted via a systematic grid search across the parameter space, as delineated in \cref{tab:space}.

For DML, we follow the methodology described in \cite{li2021decentralized}. Specifically, from the two post-distillation models, we select the model whose dataset has been utilized as the default transfer set.

For DP-KD, we employ the unsupervised version when using a public \emph{unlabeled} transfer set or the student's dataset. Conversely, when using a public \emph{labeled} transfer set, we opt for the supervised version of DP-KD, as outlined in \cref{eqn:dp-kd}. 

\begin{table}[t!]
\centering
\caption{Space of hyper-parameters.}
\begin{tabular}{lcl}
\hline
Name & Symbol & Values \\ \hline
Temperature  & $T$      & [0.1, 0.5, 1, 1.5, 2, 2.5, 3, 4, 5, 6, 7] \\
Weight       & $\alpha$ & [0.1, 0.25, 0.5, 0.75, 0.9]               \\ \hline
\end{tabular}

\label{tab:space} 
\end{table}

In our analysis, we predominantly focus on a single student and teacher setting, aligning with the assumption of a peer-to-peer collaborative system. 
Following this, in \cref{FL}, we briefly explore configurations involving multiple teachers and a single student, recognizing their potential to significantly impact centralized collaborative learning systems.
We, however, acknowledge the complexity of these multi-teacher setups and reserve their extensive exploration for future research.

\smartparagraph{Metrics.}
The primary metric of evaluation is the \emph{accuracy gain}, defined as the change in percentage points of a participant's model accuracy after KD compared to its pre-distillation accuracy. This gain may be negative, indicating a potential decline in performance due to KD.


\section{Evaluation Results}
\vspace{-2mm}
\label{sec:eval-kd}

\begin{figure*}[htbp]
  \centering
  
  \begin{subfigure}[b]{0.75\textwidth}
    \centering
    \includegraphics[width=\linewidth]{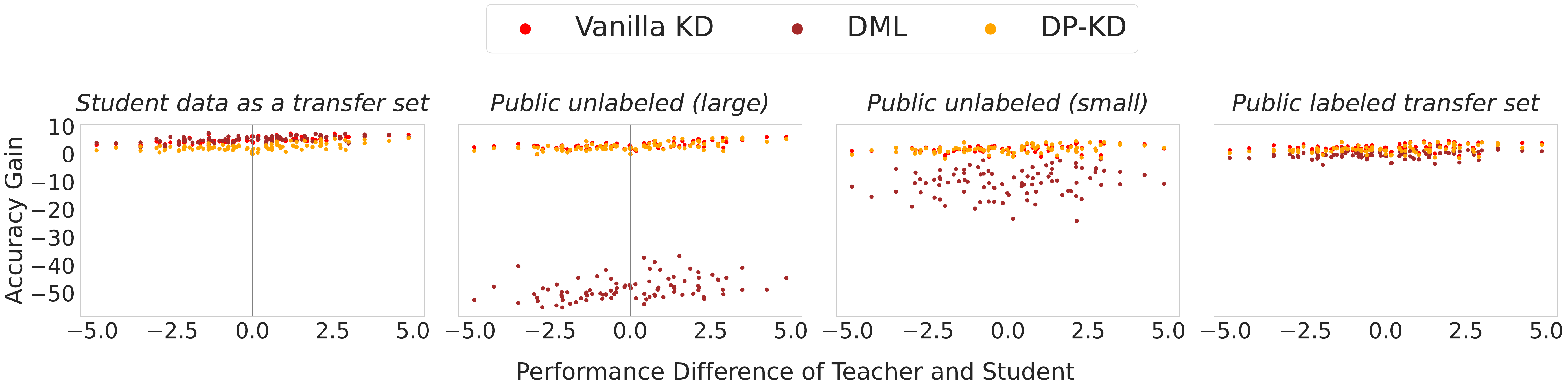}
    \caption{\small{Uniform Data Distribution}}
  \end{subfigure}
  
  \begin{subfigure}[b]{0.75\textwidth}
    \centering
    \includegraphics[width=\linewidth]{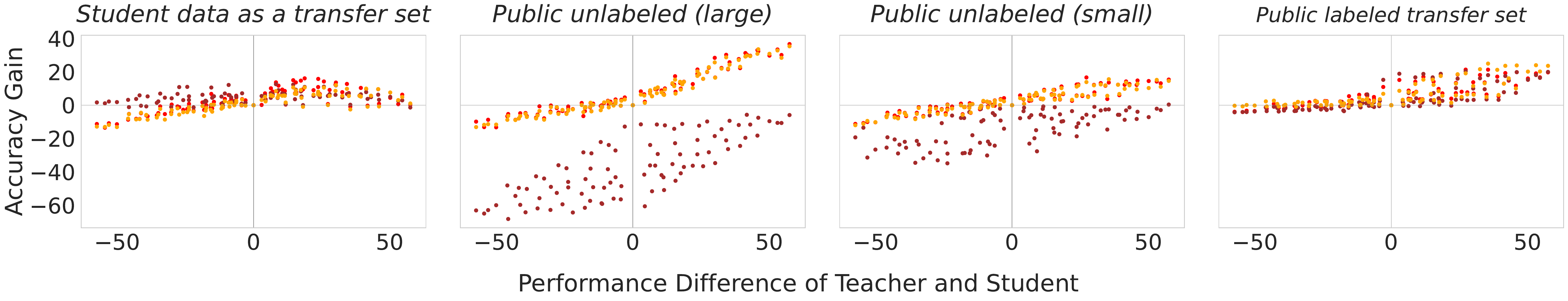}
    \caption{\small{Quantity Skew Distribution}}
  \end{subfigure}
  
  \begin{subfigure}[b]{0.75\textwidth}
    \centering
    \includegraphics[width=\linewidth]{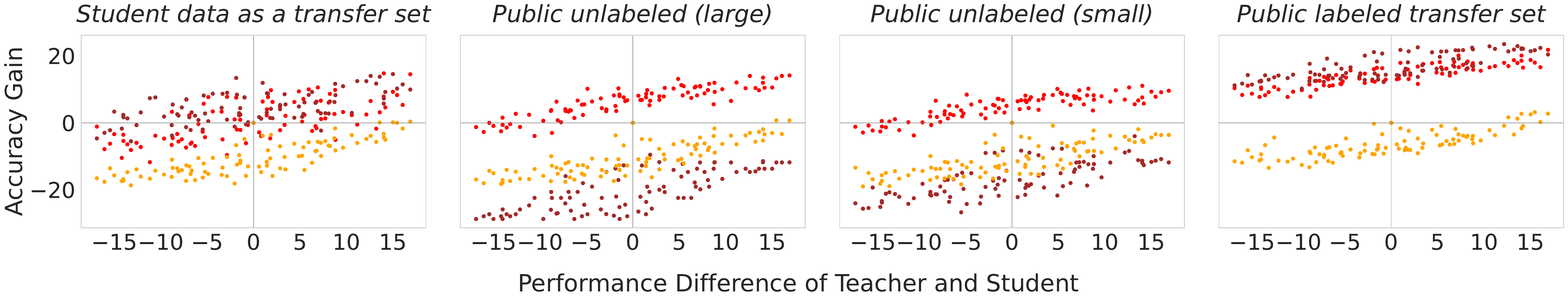}
    \caption{\small{Specialized Non-IID Distribution}}
  \end{subfigure}
  
  \begin{subfigure}[b]{0.75\textwidth}
    \centering
    \includegraphics[width=\linewidth]{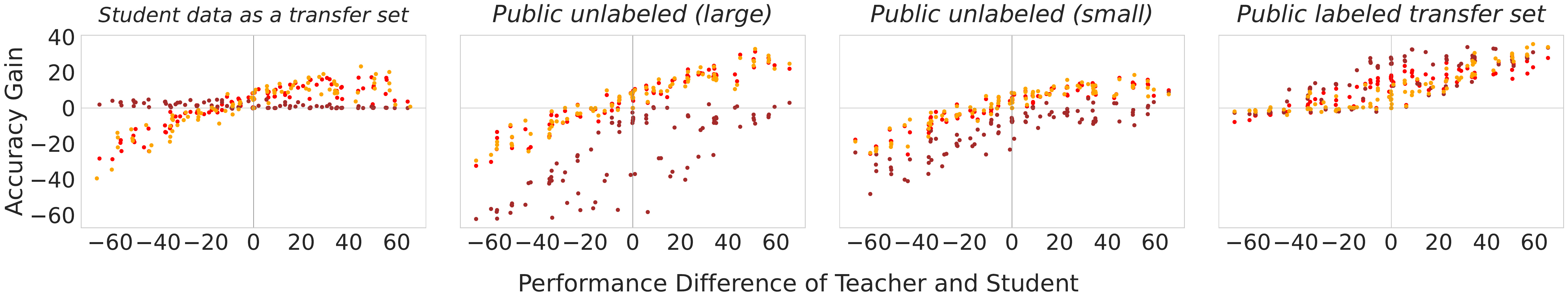}
    \caption{\small{Label Skew Random Chunks Distribution}}
  \end{subfigure}

   \begin{subfigure}[b]{0.75\textwidth}
    \centering
    \includegraphics[width=\linewidth]{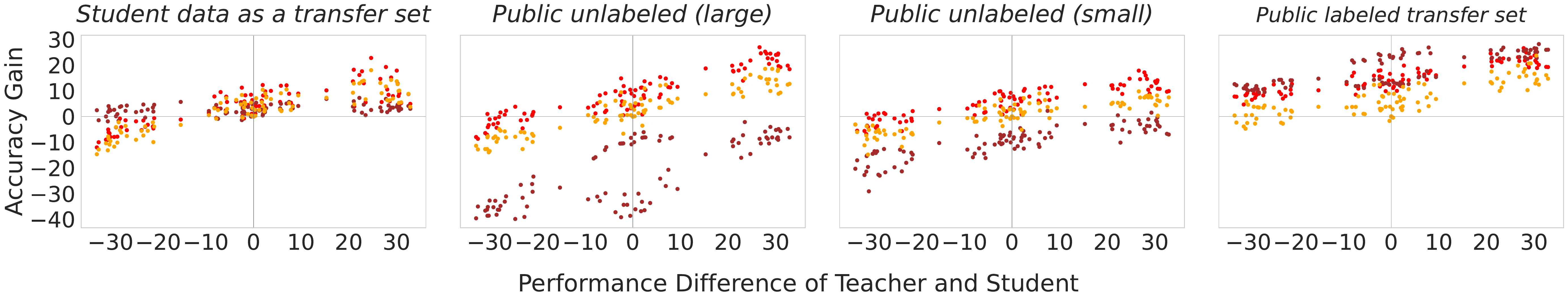}
    \caption{\small{Label Skew Dirichlet Distribution}}
  \end{subfigure}
  \caption{Evaluation of KD techniques across various data partitioning strategies and transfer set options.}
  \label{fig:data_partitioning_strategies}
\end{figure*}

We compare various KD approaches and demonstrate their performance across different data distribution scenarios and transfer set configurations.
\cref{fig:data_partitioning_strategies} illustrates the results. Each subplot of the figure is a scatter plot showing knowledge transfer across all 90 participant pairs. Each data point in these subplots corresponds to a distinct knowledge transfer instance between two models. The x-axis denotes the performance difference between the teacher and the student models (quantified as teacher accuracy minus student accuracy), while the y-axis value reflects the accuracy gain achieved via KD. 

Teachers with notably higher accuracy than their corresponding students are categorized as \emph{strong}, while those with lower accuracy relative to the students are designated as \emph{weak}.
Instances of KD that result in improved accuracy are designated as \emph{positive knowledge transfers}. 

This comparative analysis includes transfer sets of varying sizes to determine their impact on the efficacy of KD. Specifically, the ``Public unlabeled (large)'' transfer set comprises 5,000 samples, and the ``Public unlabeled (small)'' and the ``Public labeled transfer set'' each contain 500 samples, randomly sampled from the original dataset before partitioning. The ``Student data as a transfer set'' varies in size, corresponding to the student dataset. 
\vspace{-3mm}

We now analyze our main findings from this comparison.
First, we contrast the performance of DML and DP-KD with that of Vanilla KD, representing the baseline approach. Then, we illustrate the impact of hyperparameter tuning by comparing Tuned KD against other approaches.

\subsection{Different transfer set options}
\vspace{-1mm}
\begin{findingbox}
\textbf{Finding (1):} Increasing the number of samples in the transfer set enhances the effectiveness of KD, leading to more pronounced positive knowledge transfers.
\end{findingbox}
\vspace{-1mm}
Our analysis reveals a proportional relationship between the size of the transfer set and the effectiveness of knowledge distillation. As demonstrated in \cref{fig:data_partitioning_strategies}, expanding the transfer set size consistently improves the outcomes of knowledge transfer. This improvement is particularly noticeable when comparing positive knowledge transfers in columns 2 and 3 of the figure.
This enhancement in KD effectiveness with larger transfer sets can be attributed to the increased diversity and representativeness of the sample distribution. A larger set provides more comprehensive coverage of the input space, allowing the student model to learn a more generalized data representation. 
This principle aligns with statistical learning theory, which posits that larger datasets can reduce the variance of estimators and improve model generalization on unseen data \cite{vapnik2013nature}. Furthermore, supporting research \cite{phuong2019towards} indicates that an expanded transfer set can significantly diminish the student classifier's loss, thereby aligning the student's parameters more closely with the teacher model.

It is worth noting that a larger sample size in the transfer set may also lead to more pronounced negative knowledge transfers under certain conditions, as observed with the DML approach in \cref{fig:data_partitioning_strategies} (compare columns 2 and 3). As discussed next, the DML approach is not recommended without a student dataset or another public labeled transfer set.

\begin{findingbox}
\textbf{Finding (2):} The DML approach performs poorly in unsupervised transfer set settings.
\end{findingbox}
\vspace{-2mm}
Our findings indicate that DML generally outperforms Vanilla KD and DP-KD in scenarios involving weaker teachers, particularly in scenarios where `Student data' and `Public labelled data' are used as transfer sets. On the other hand, in the absence of labeled data, its performance significantly diminishes. 
In such unsupervised settings, DML can reinforce existing biases or errors between the models, as there is no ground truth to guide the correction of predictions. This issue is further exacerbated by the peer-to-peer learning structure of DML, where two models continuously adjust their parameters based on potentially flawed outputs from each other, leading to compounded inaccuracies without proper error correction mechanisms.
This often results in negative knowledge transfers, as observed in columns 2 and 3 of \cref{fig:data_partitioning_strategies}. This pattern cautions against employing DML when neither student data nor a labeled transfer set is accessible.

\vspace{-1mm}
\subsection{DML and DP-KD vs. Vanilla KD}

\begin{findingbox}
\textbf{Finding (3):} Vanilla KD demonstrates similar or superior performance compared to DP-KD across various data distribution scenarios.
\end{findingbox}
\vspace{-1mm}
\cref{fig:data_partitioning_strategies} provides a detailed comparative analysis underscoring this finding. In Uniform Data Distribution, Quantity Skew Distribution, and Label Skew Random Chunks Distribution scenarios, Vanilla KD's performance is comparable to that of DP-KD. However, it notably excels in the Label Skew Dirichlet Distribution scenario and demonstrates significant superiority in the Specialized Non-IID Distribution context, across all tested transfer set options. This pattern of performance highlights the robustness and adaptability of Vanilla KD, particularly in complex data distribution scenarios and when paired with teachers of comparable or superior strength.

\begin{findingbox}
\textbf{Finding (4):} DP-KD performs poorly in the `Specialized Non-IID Distribution' scenario.
\end{findingbox}
\vspace{-2mm}
DP-KD's underperformance in the Specialized Non-IID Distribution scenario, which features significant class imbalances, highlights its challenges in handling skewed data distributions. Here, even though the teacher model has encountered samples from all classes, its potential overconfidence in over-represented classes can lead to misleading guidance for the student model. This misalignment often results in the reinforcement of errors rather than corrective learning. This underscores the importance of considering class balance in the teacher's training dataset for effective DP-KD application.

\begin{findingbox}
\textbf{Finding (5):} 
There is a notable susceptibility in employing Vanilla KD and DP-KD methods with weak teachers, especially when using student data as the transfer set.
\end{findingbox}
\vspace{-2mm}
The limited effectiveness of Vanilla KD and DP-KD when paired with weaker teachers, particularly when using student data as the transfer set, underscores a significant drawback in these methods under specific circumstances. The performance decline can be theoretically linked to the weaker teachers' insufficient guidance, which inadequately boosts the student model's learning, especially when leveraging potentially less diverse student data

DML demonstrates an ability to extract useful knowledge from these interactions, even with weaker teacher models, as evidenced in \cref{fig:data_partitioning_strategies}, particularly in column 1. This indicates DML’s ability to mutually benefit participating models, effectively utilizing their collective insights despite individual limitations. However, Vanilla KD tends to surpass DML in scenarios involving stronger teachers. This indicates that the method choice is dependent on the teacher models' strength, i.e., DML is preferred when teacher models are less powerful, and Vanilla KD is preferred when teacher models are robust.

\subsection{Optimizing Hyperparameters: Tuned KD}

\begin{figure*}[htbp]
  \centering
  \begin{tabularx}{0.85\textwidth}{XXXX}
    
    \begin{adjustbox}{height=1.3in, keepaspectratio} 
    
    \includegraphics[width=\linewidth]{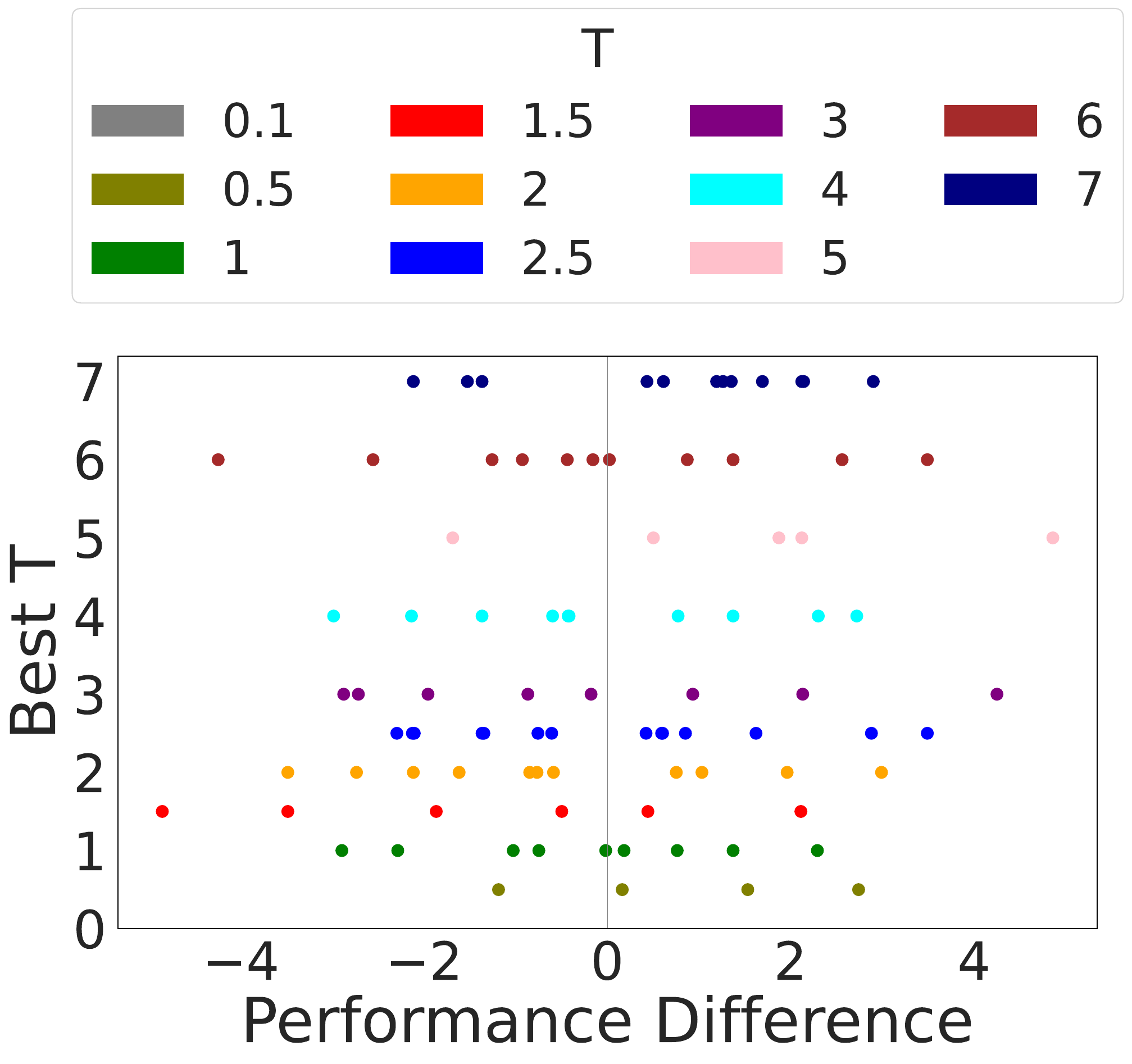} \end{adjustbox}&
    \begin{adjustbox}{height=1.22in, keepaspectratio} 
    \includegraphics[width=\linewidth]{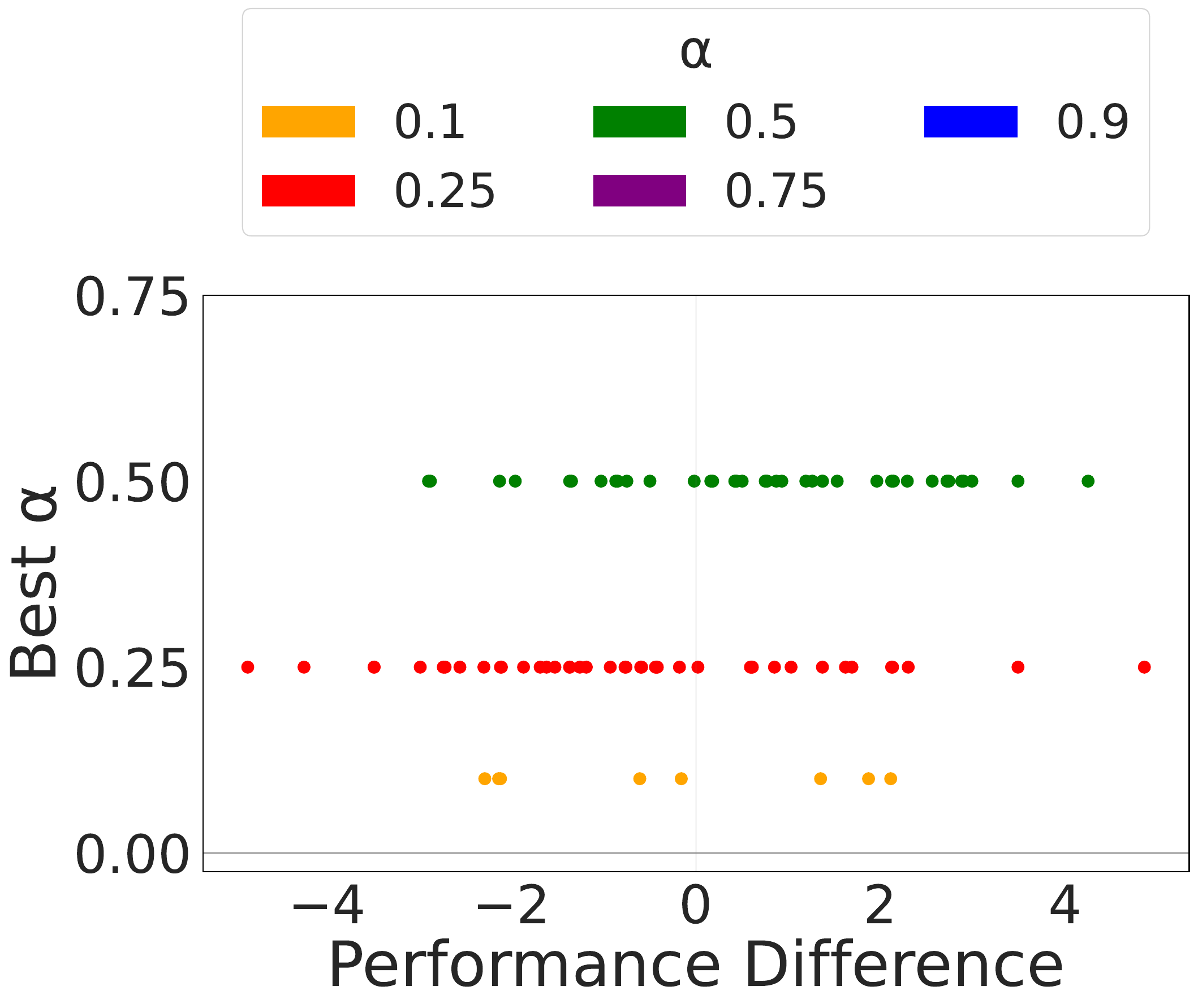} \end{adjustbox}&
    \begin{adjustbox}{height=1.3in, keepaspectratio}
    \hspace{6mm} 
    \includegraphics[width=\linewidth]{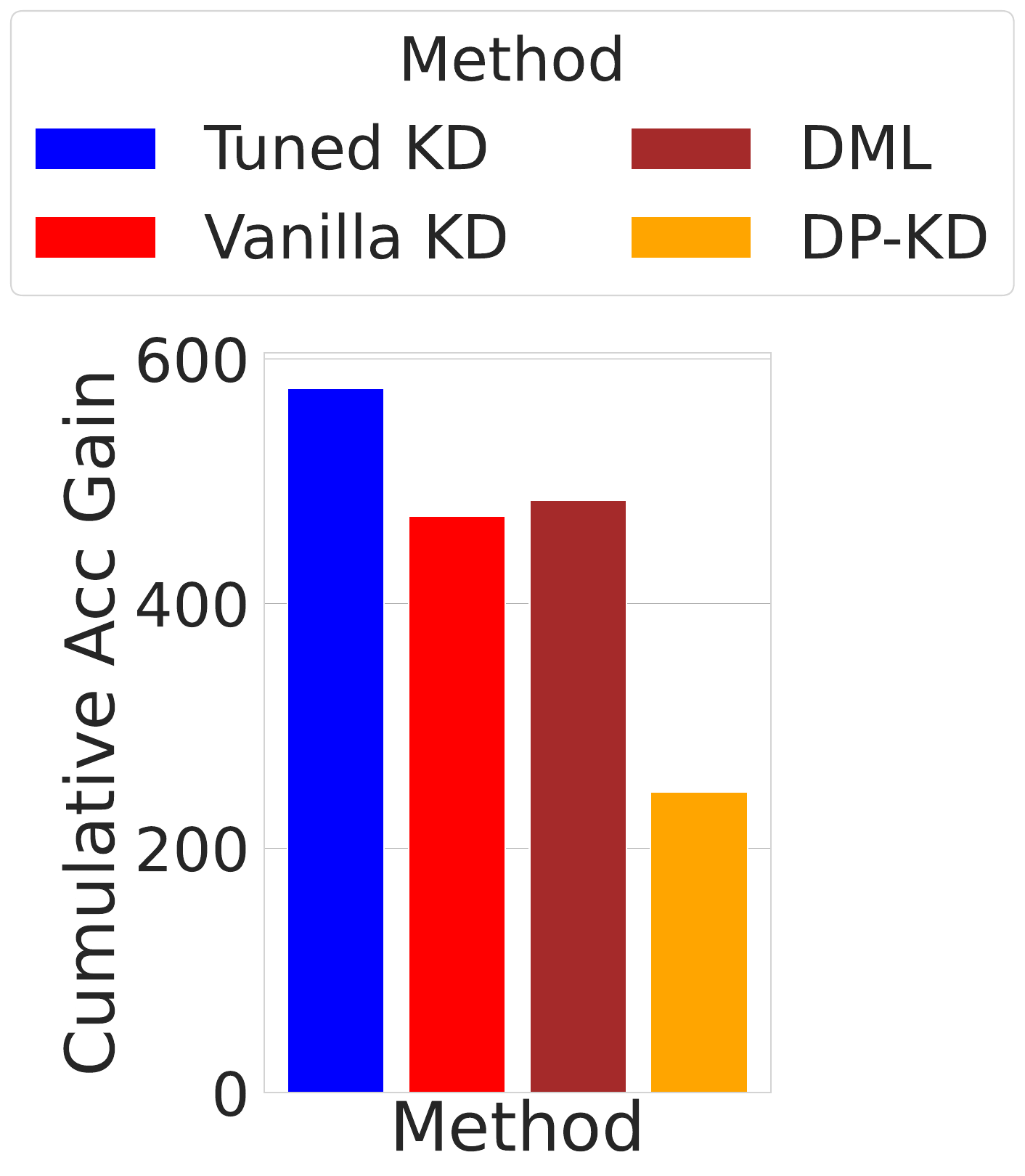} \end{adjustbox}&
    \begin{adjustbox}{height=1.18in, keepaspectratio} \hspace{1mm}
    \includegraphics[width=\linewidth]{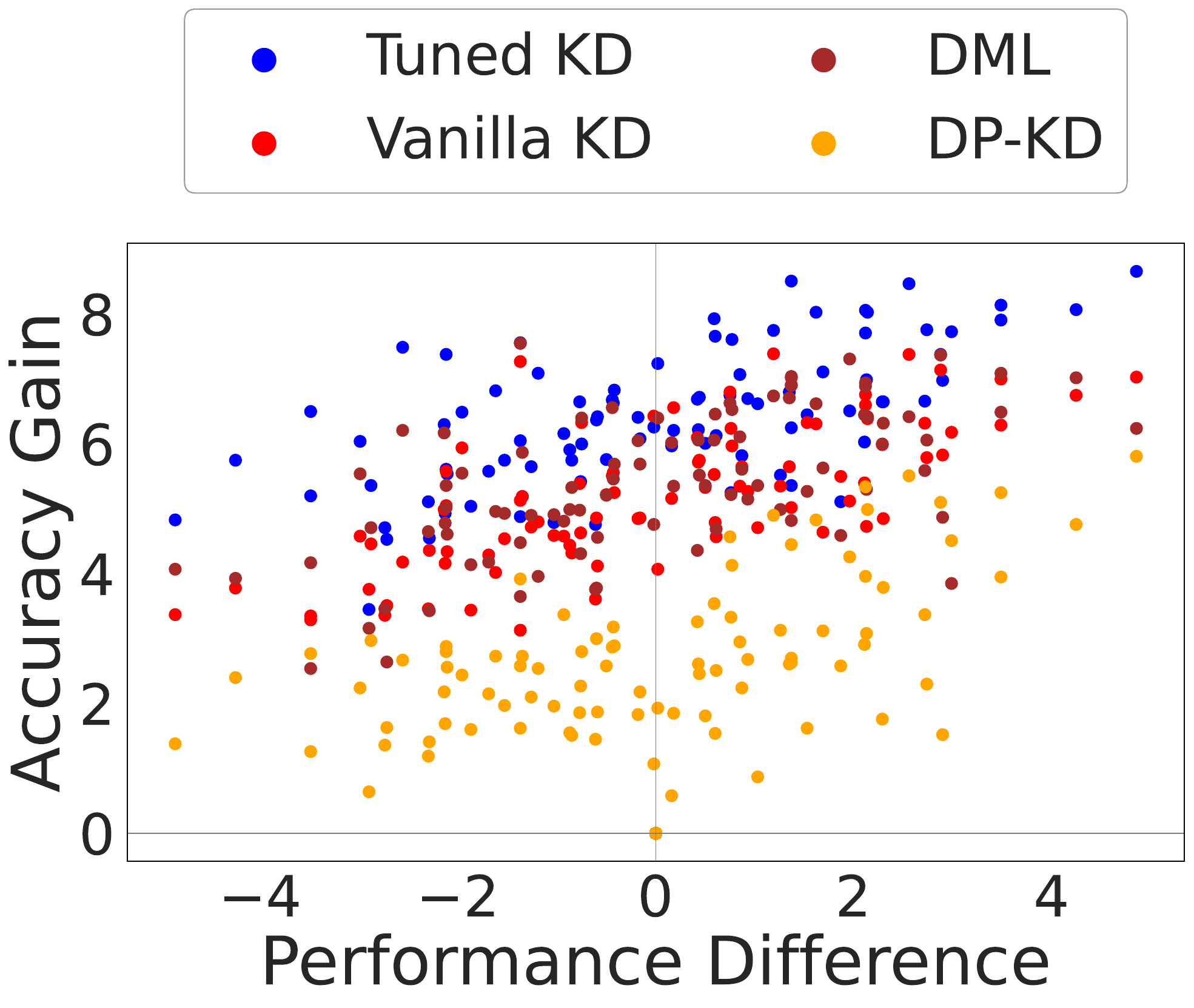} \end{adjustbox}\\
    \multicolumn{4}{c}{\textbf{\small{(a) Uniform Data Distribution}}} \\

    \begin{adjustbox}{height=0.9in, keepaspectratio}
    \includegraphics[width=\linewidth]{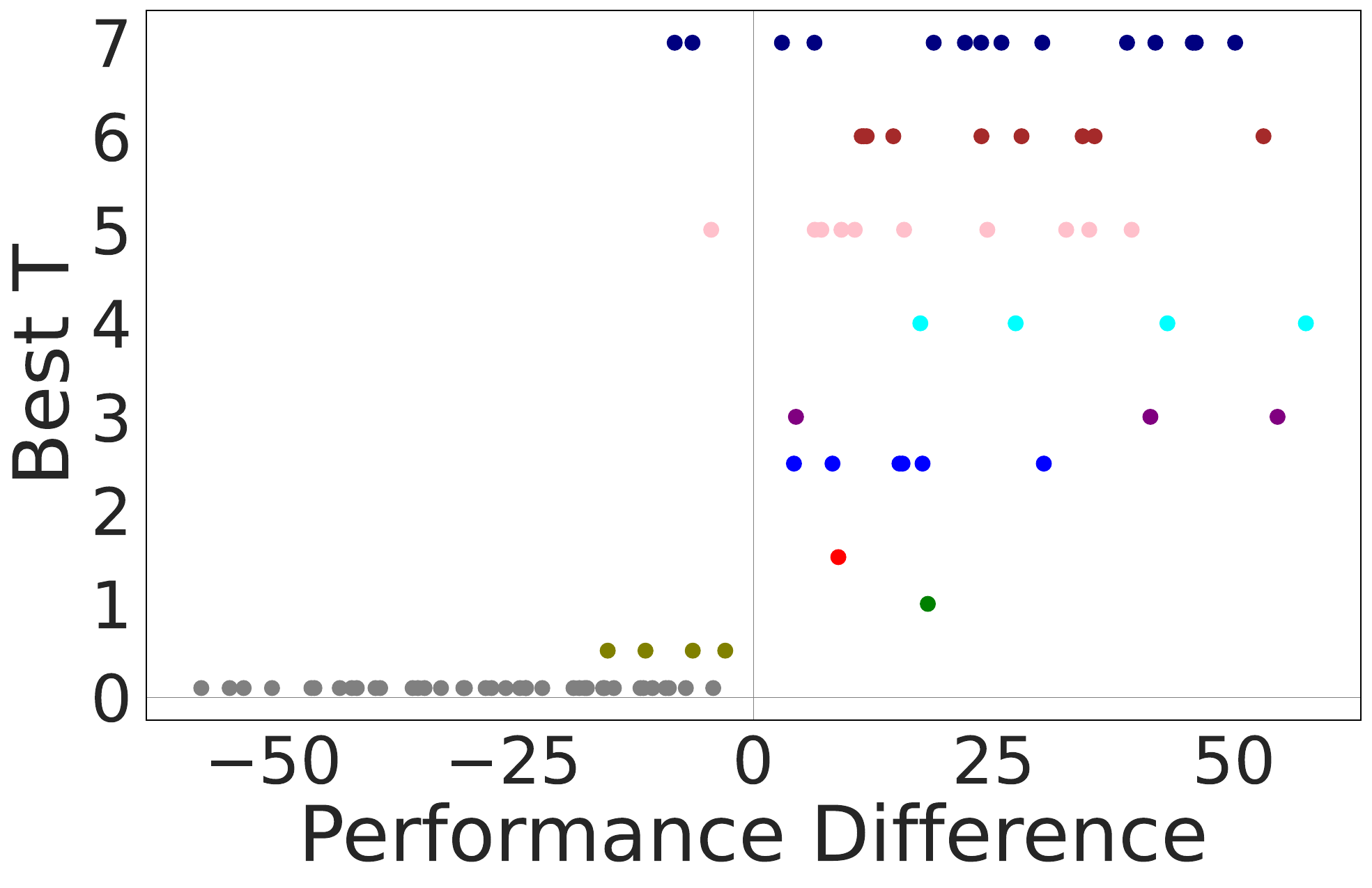} \end{adjustbox}&
    \begin{adjustbox}{height=0.9in, keepaspectratio}
    \includegraphics[width=\linewidth]{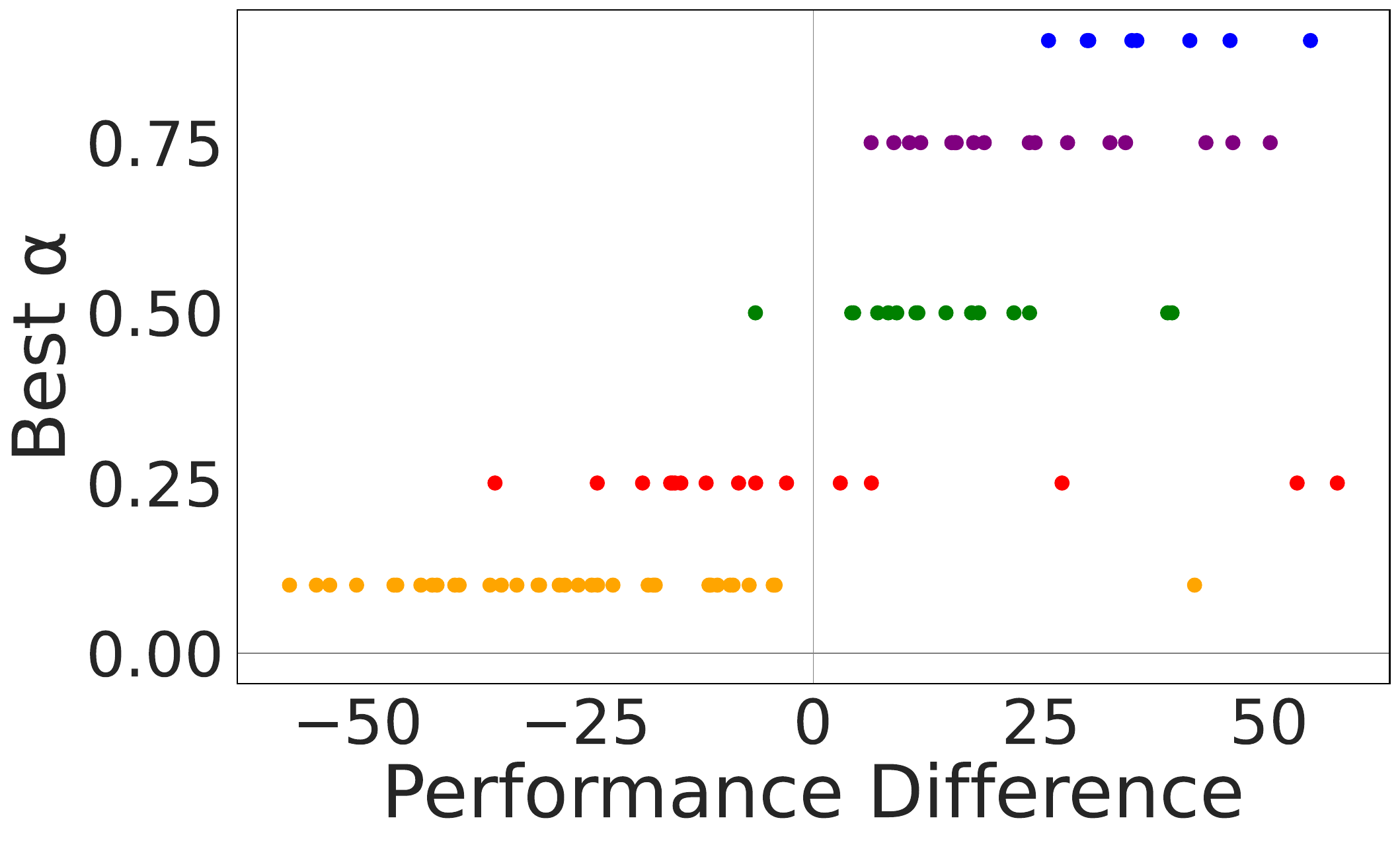} \end{adjustbox}&
    \begin{adjustbox}{height=0.9in, keepaspectratio}
    \hspace{13mm} 
    \includegraphics[width=\linewidth]{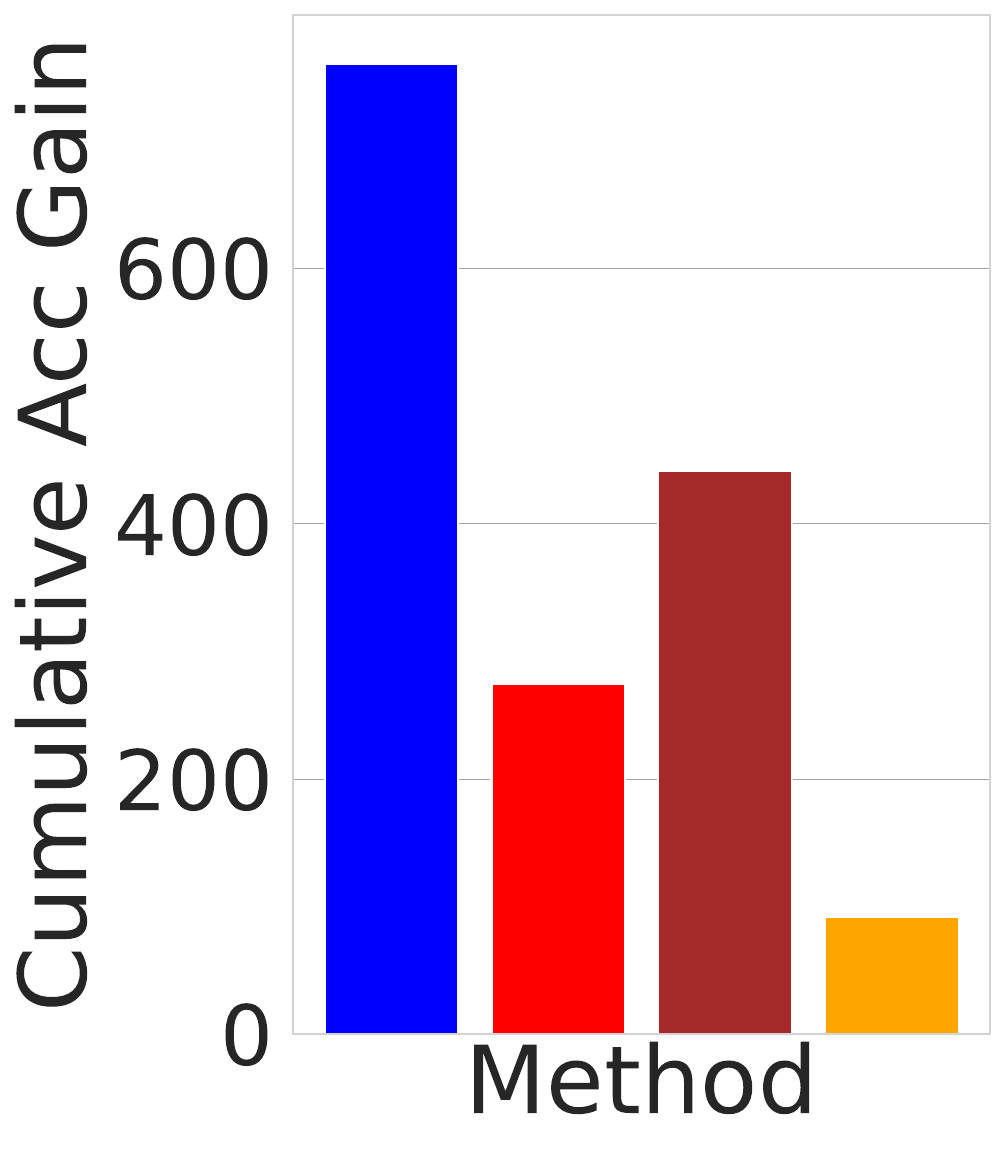} \end{adjustbox}&
    \begin{adjustbox}{height=0.9in, keepaspectratio}
    \includegraphics[width=\linewidth]{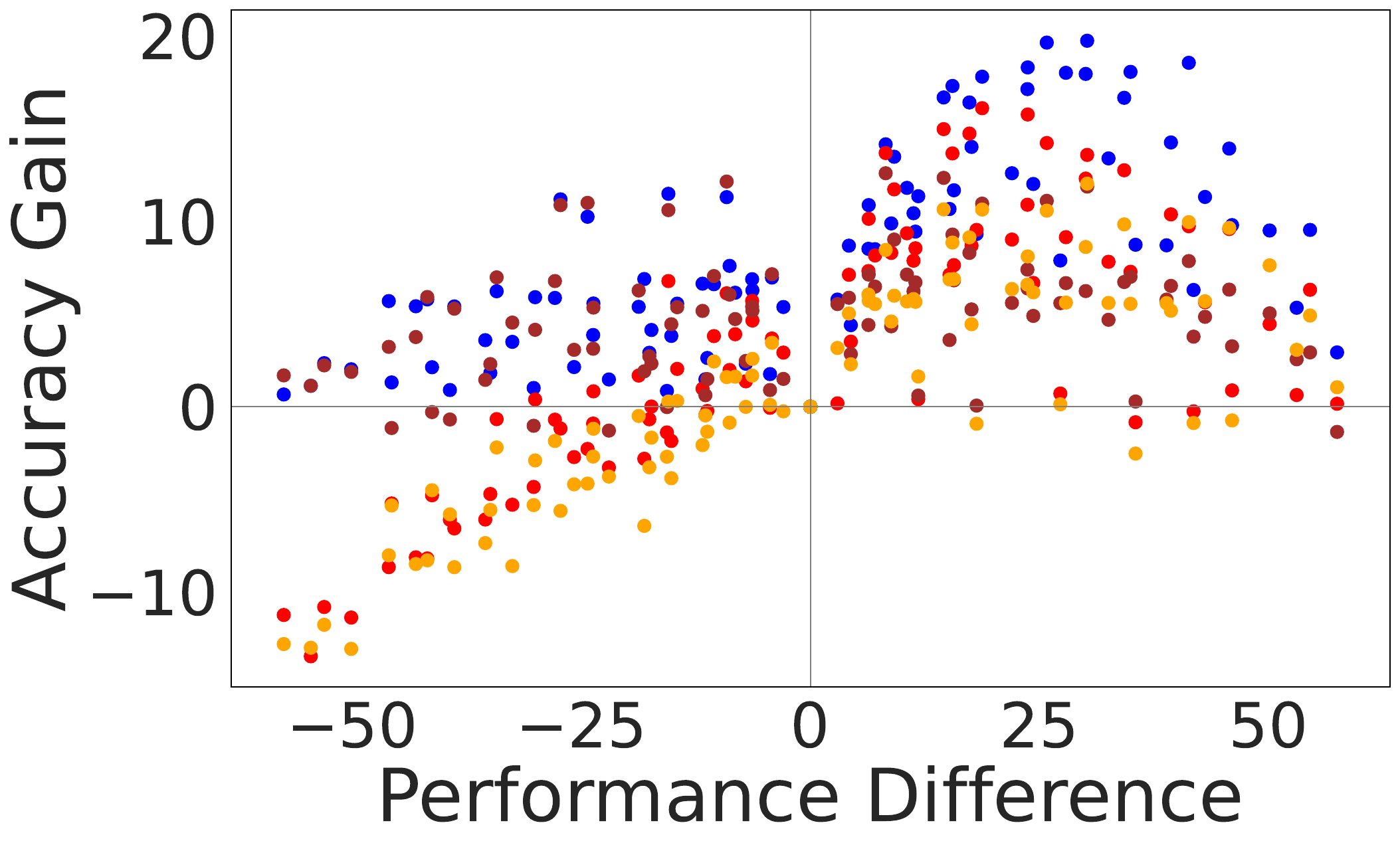} \end{adjustbox}\\
    \multicolumn{4}{c}{\textbf{\small{(b) Quantity Skew Distribution}}} \\

    \begin{adjustbox}{height=0.9in, keepaspectratio}
    \includegraphics[width=\linewidth]{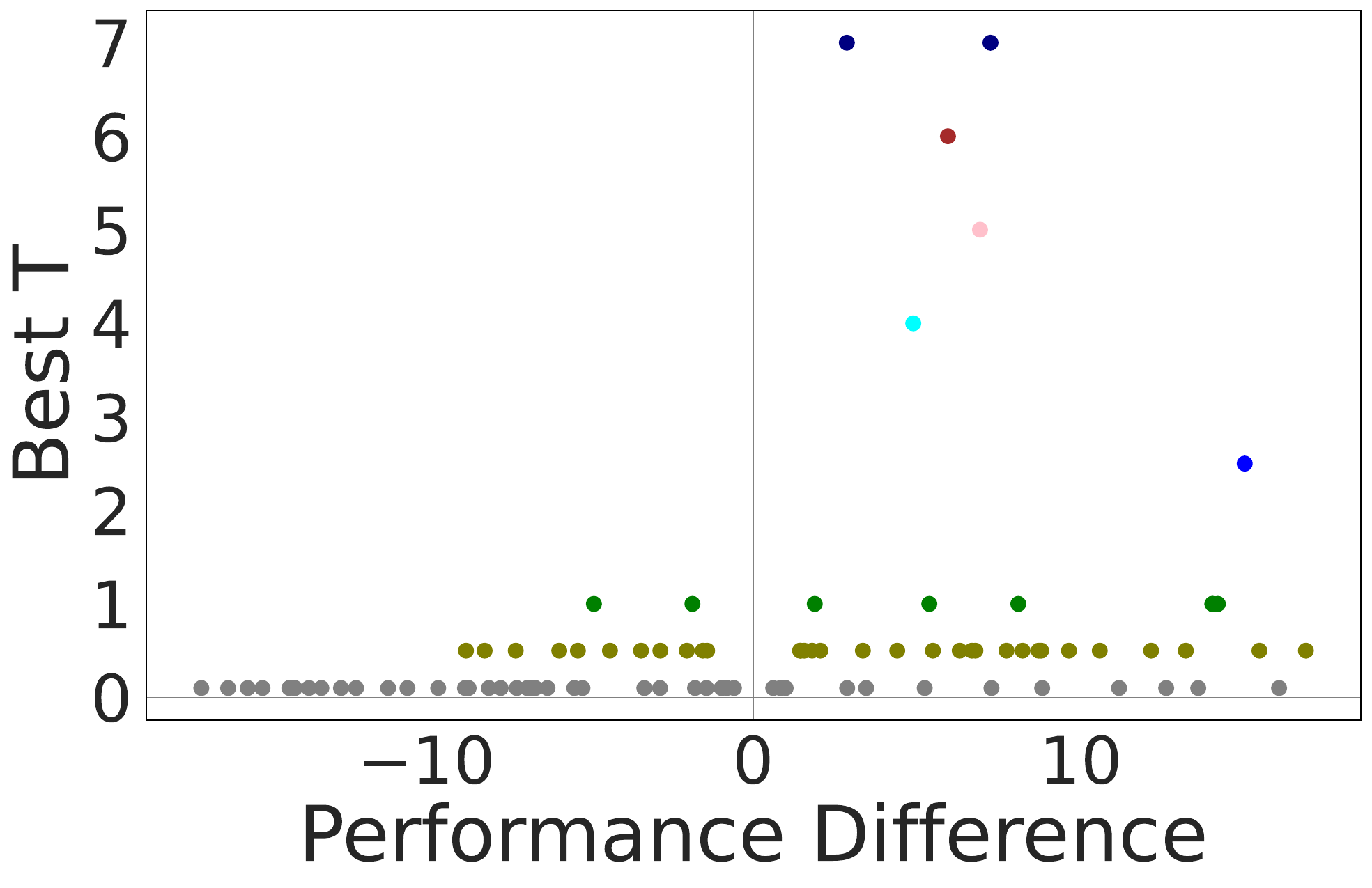} \end{adjustbox}&
    \begin{adjustbox}{height=0.9in, keepaspectratio}
    \includegraphics[width=\linewidth]{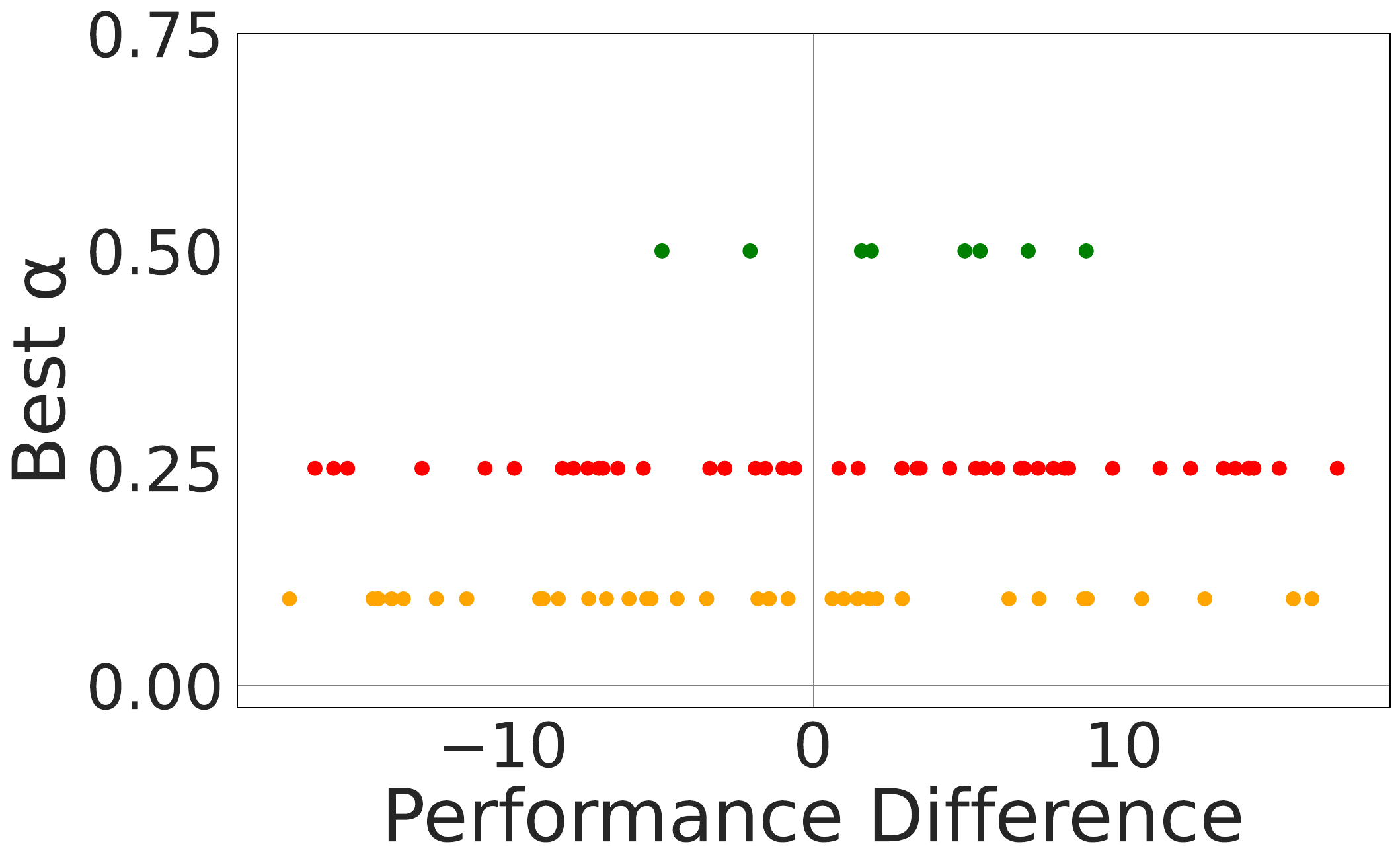} \end{adjustbox}&
    \begin{adjustbox}{height=0.9in, keepaspectratio}
    \hspace{9mm} 
    \includegraphics[width=\linewidth]{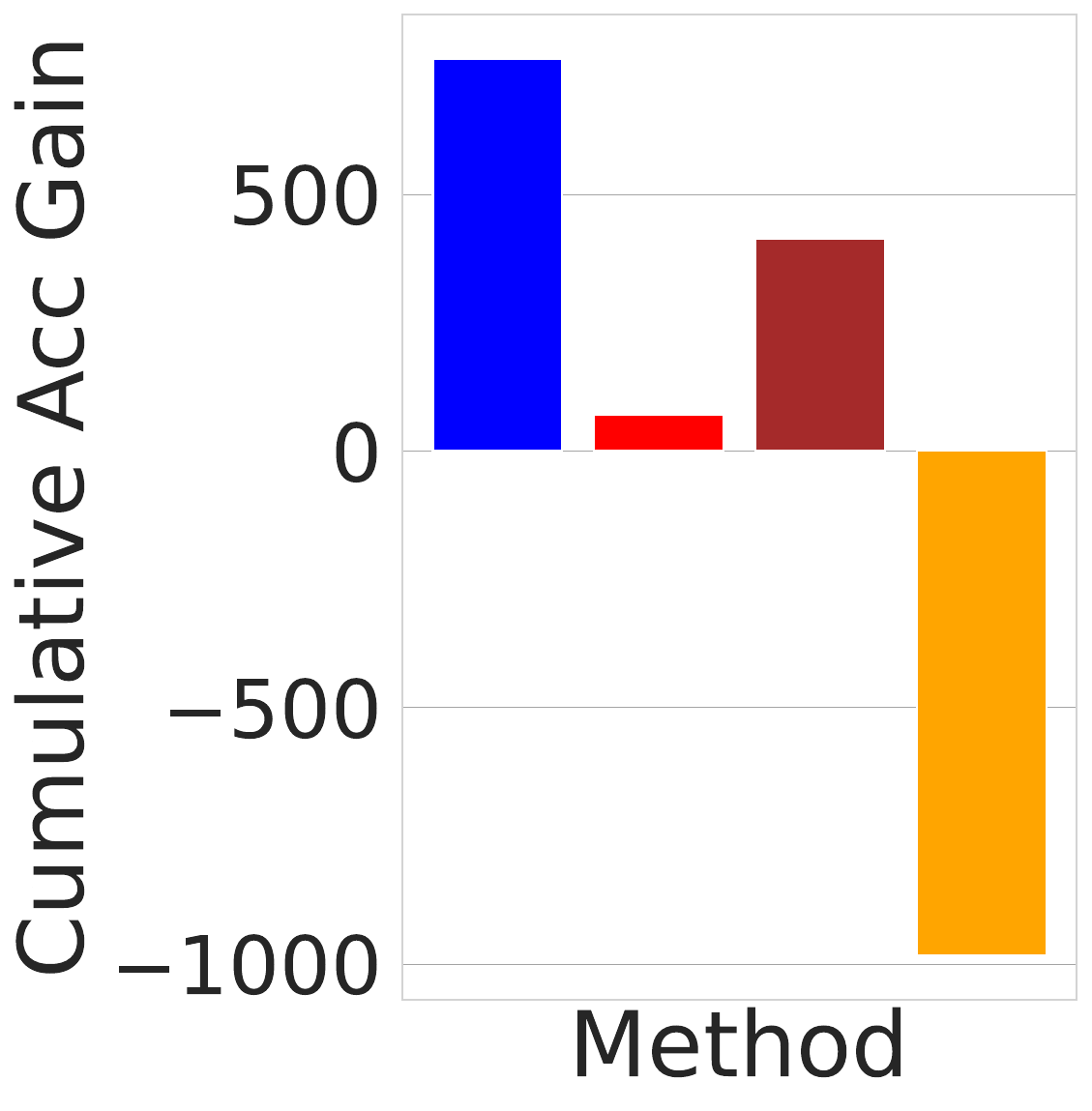} \end{adjustbox}&
    \begin{adjustbox}{height=0.9in, keepaspectratio}
    \includegraphics[width=\linewidth]{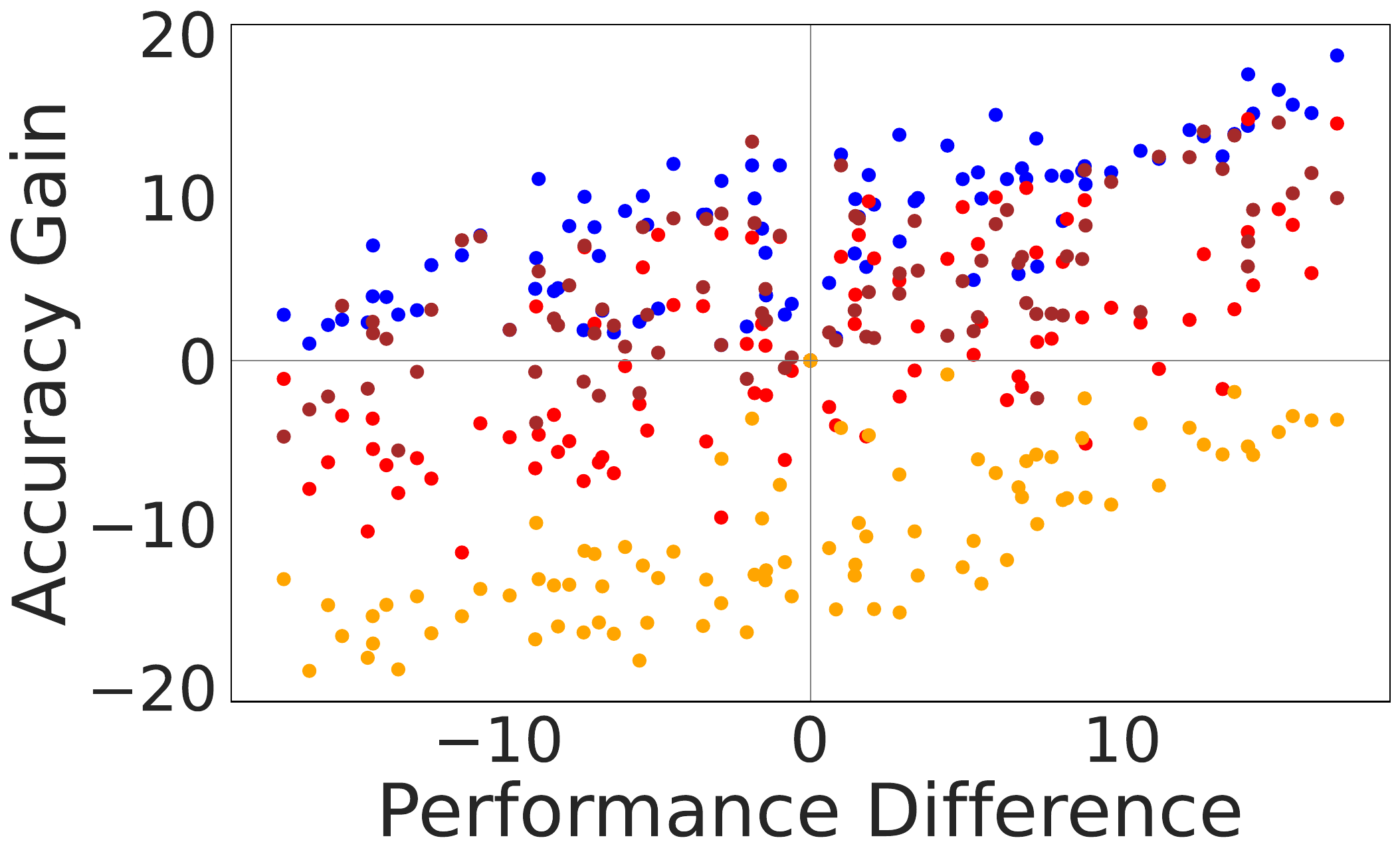} \end{adjustbox}\\
    \multicolumn{4}{c}{\textbf{\small{(c) Specialized Non-IID Distribution}}} \\

    \begin{adjustbox}{height=0.9in, keepaspectratio}
    \includegraphics[width=\linewidth]{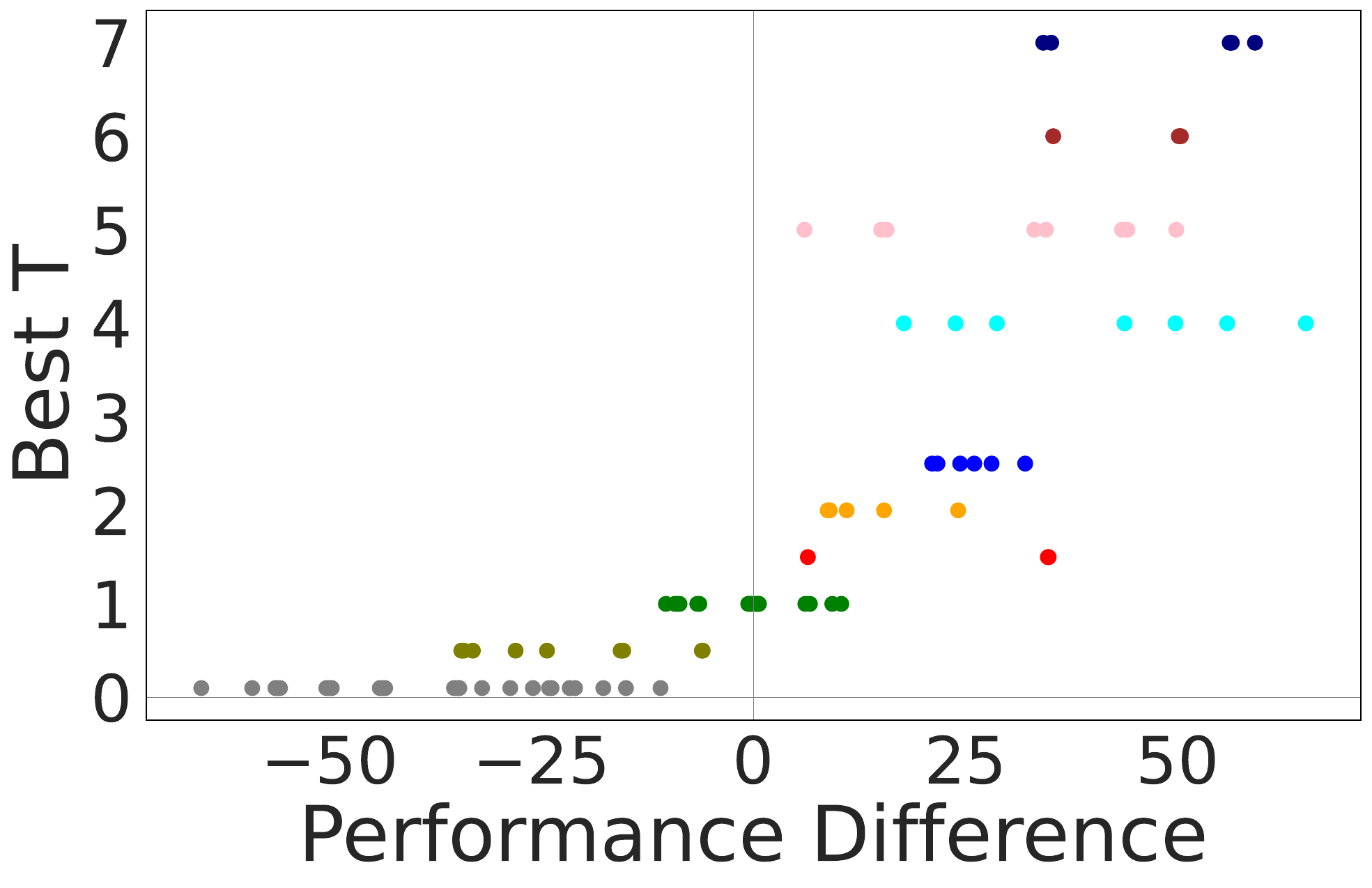} \end{adjustbox}&
    \begin{adjustbox}{height=0.9in, keepaspectratio}
    \includegraphics[width=\linewidth]{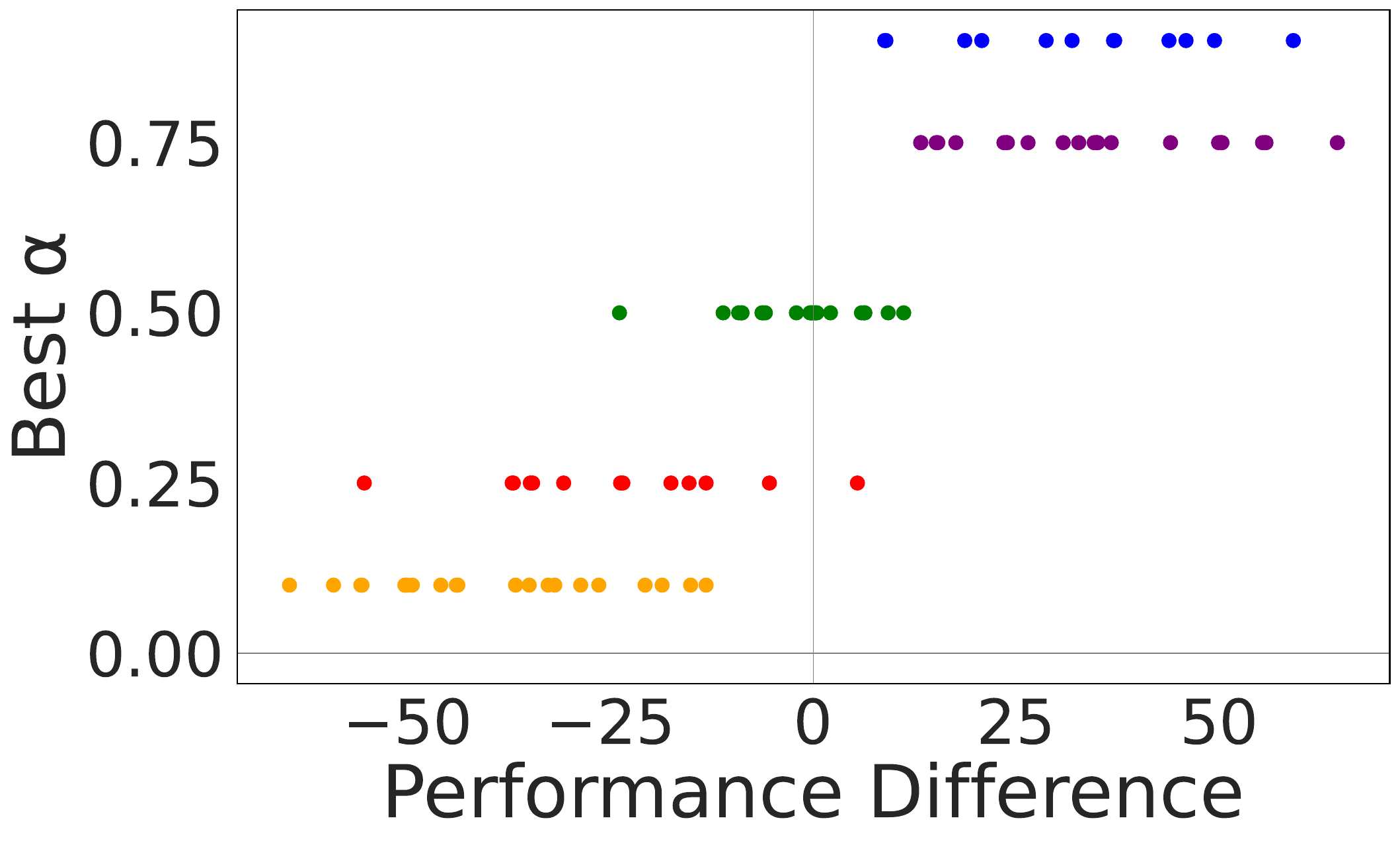} \end{adjustbox}&
    \begin{adjustbox}{height=0.9in, keepaspectratio}
    \hspace{13mm} 
    \includegraphics[width=\linewidth]{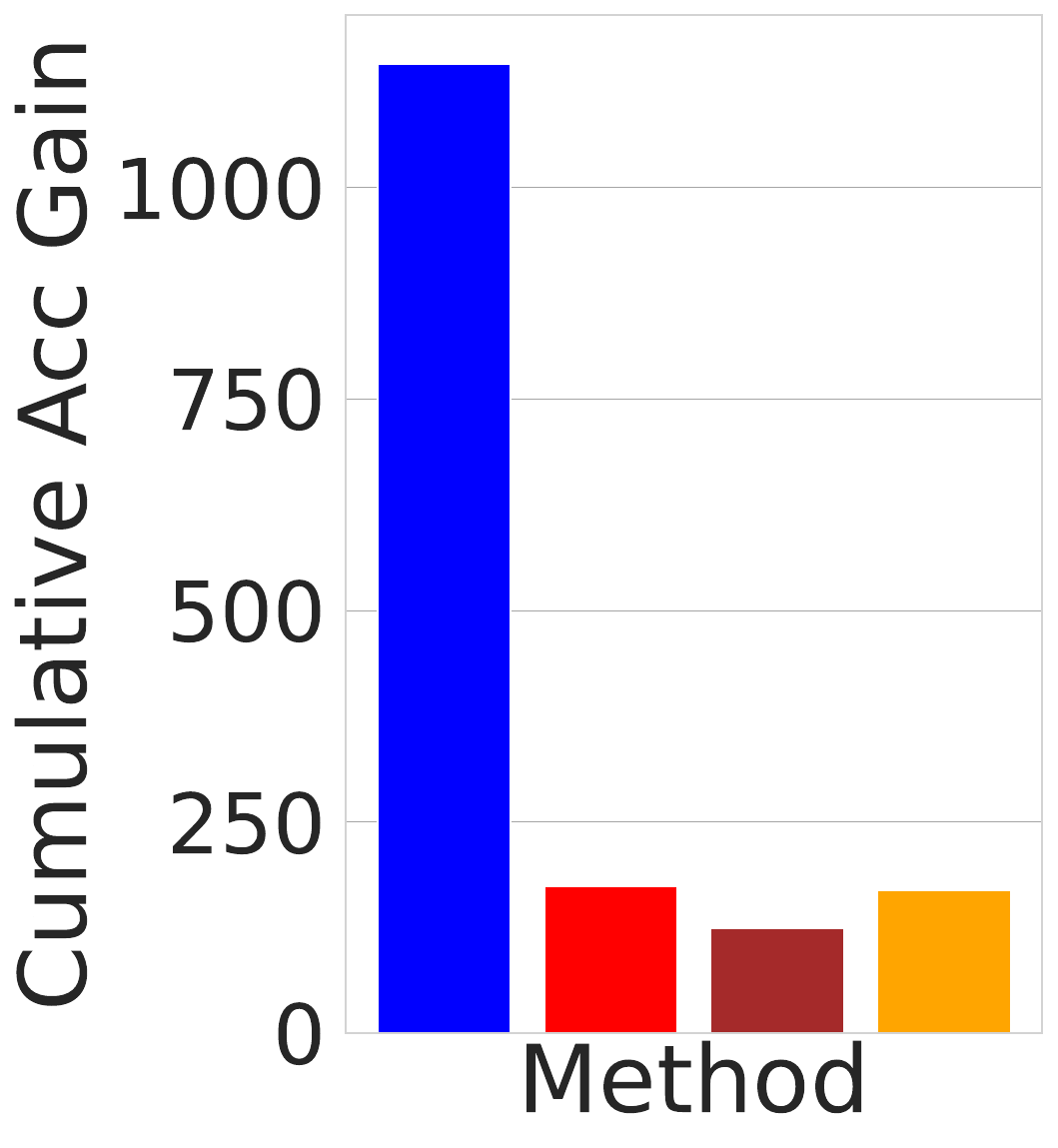} \end{adjustbox}&
    \begin{adjustbox}{height=0.9in, keepaspectratio}
    \includegraphics[width=\linewidth]{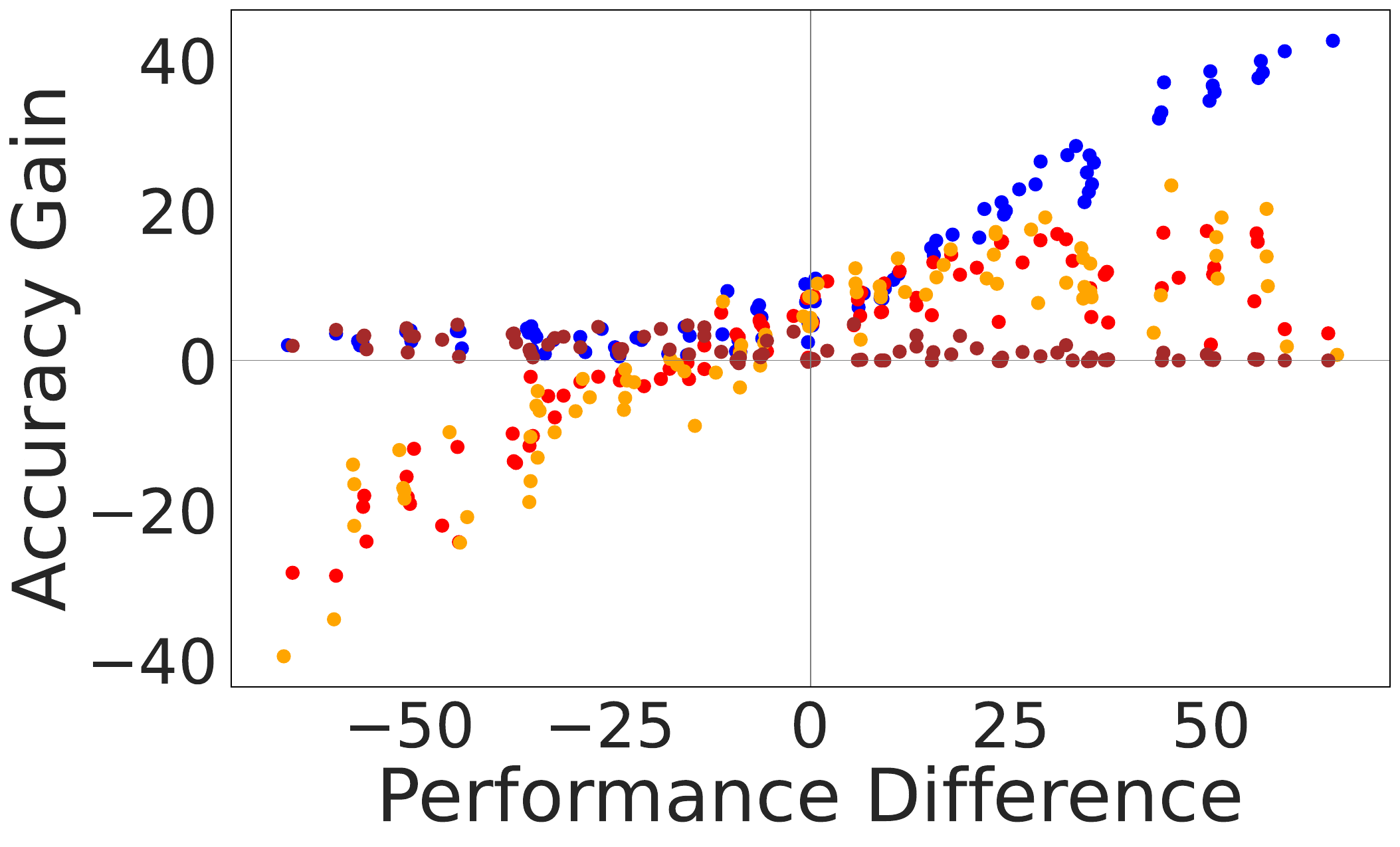} \end{adjustbox}\\
    \multicolumn{4}{c}{\textbf{\small{(d) Label Skew Random Chunks Distribution}}} \\

    \begin{adjustbox}{height=0.9in, keepaspectratio}
    \includegraphics[width=\linewidth]{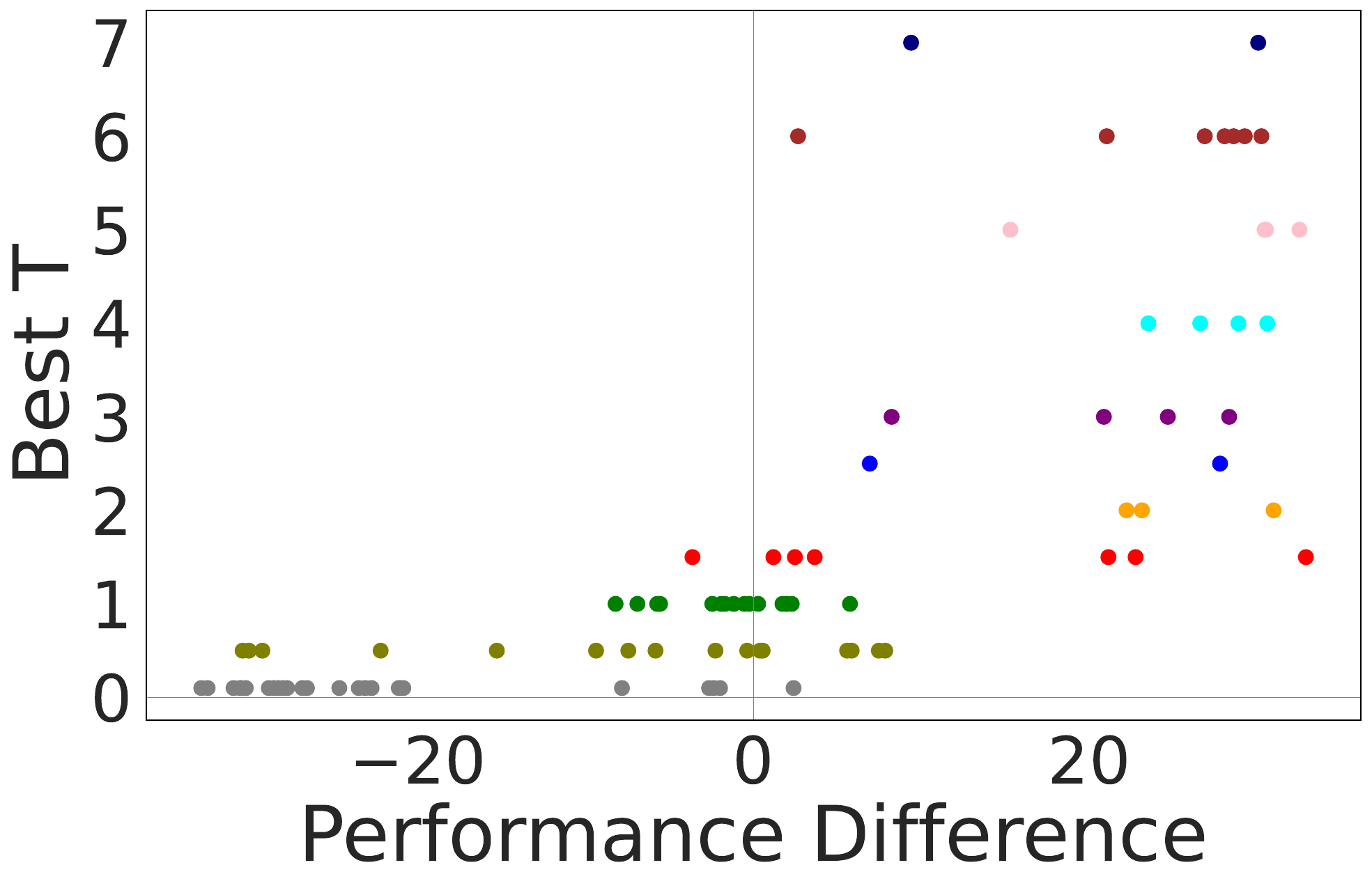} \end{adjustbox}&
    \begin{adjustbox}{height=0.9in, keepaspectratio}
    \includegraphics[width=\linewidth]{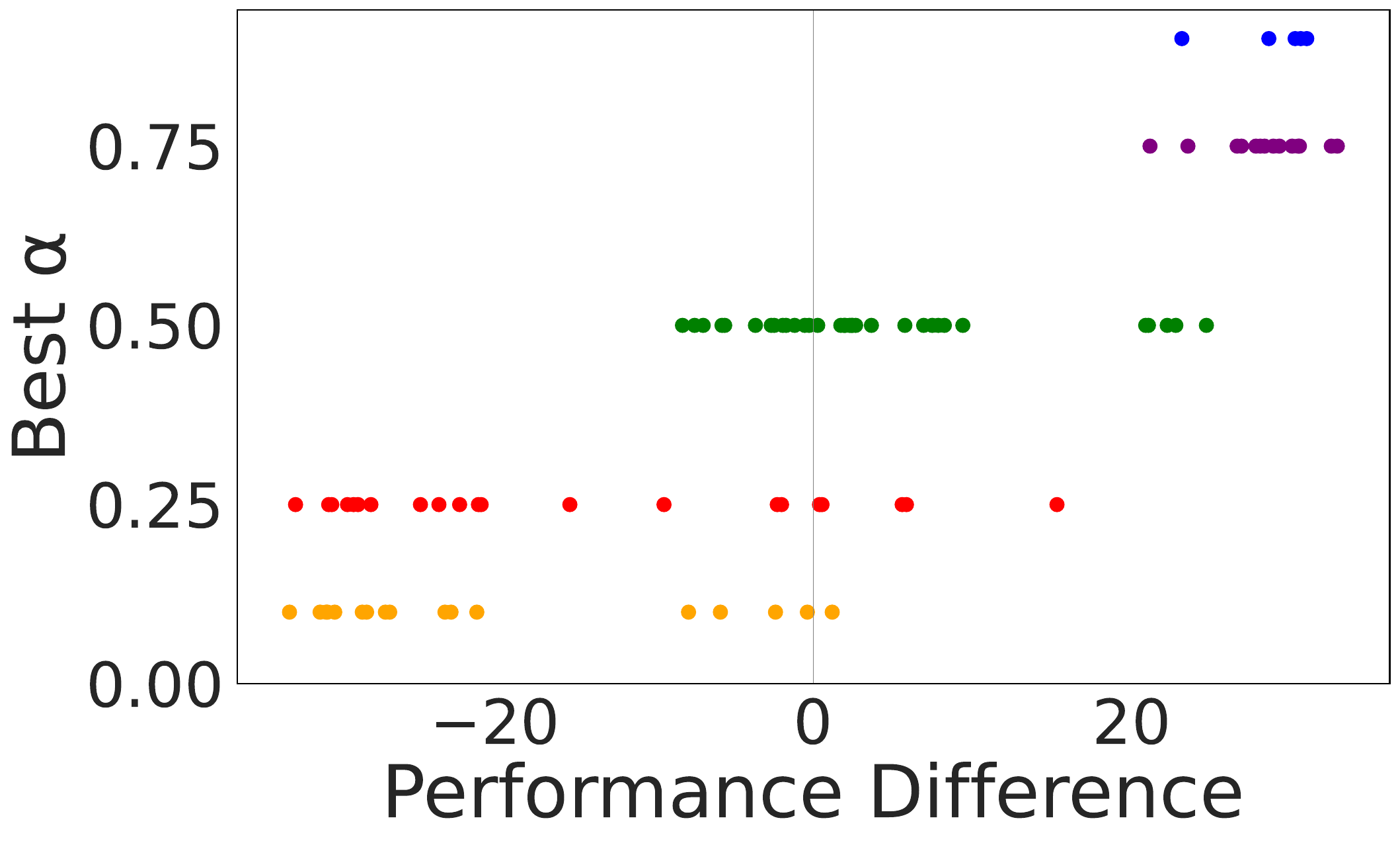} \end{adjustbox}&
    \begin{adjustbox}{height=0.9in, keepaspectratio}
    \hspace{15mm} 
    \includegraphics[width=\linewidth]{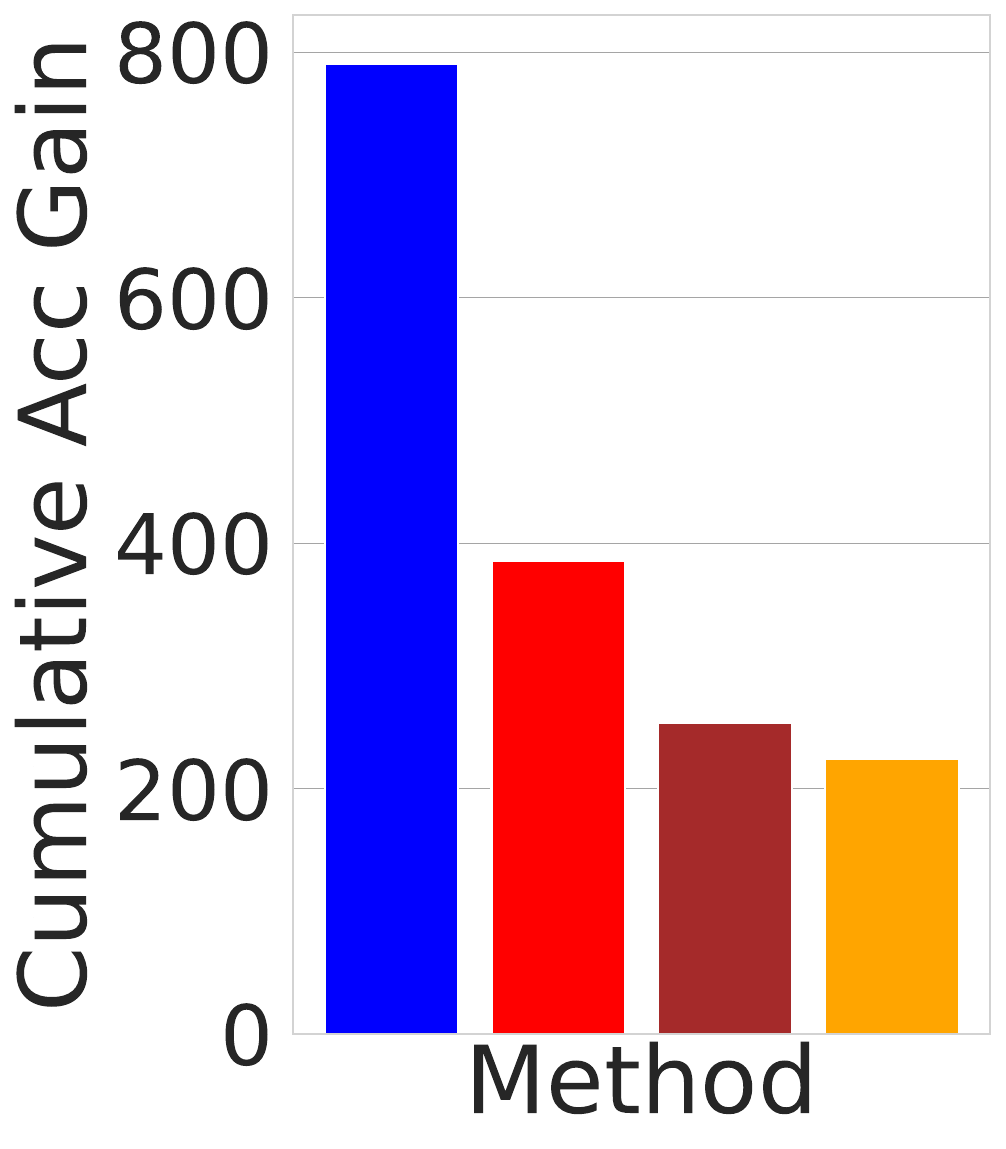} \end{adjustbox}&
    \begin{adjustbox}{height=0.9in, keepaspectratio}
    \includegraphics[width=\linewidth]{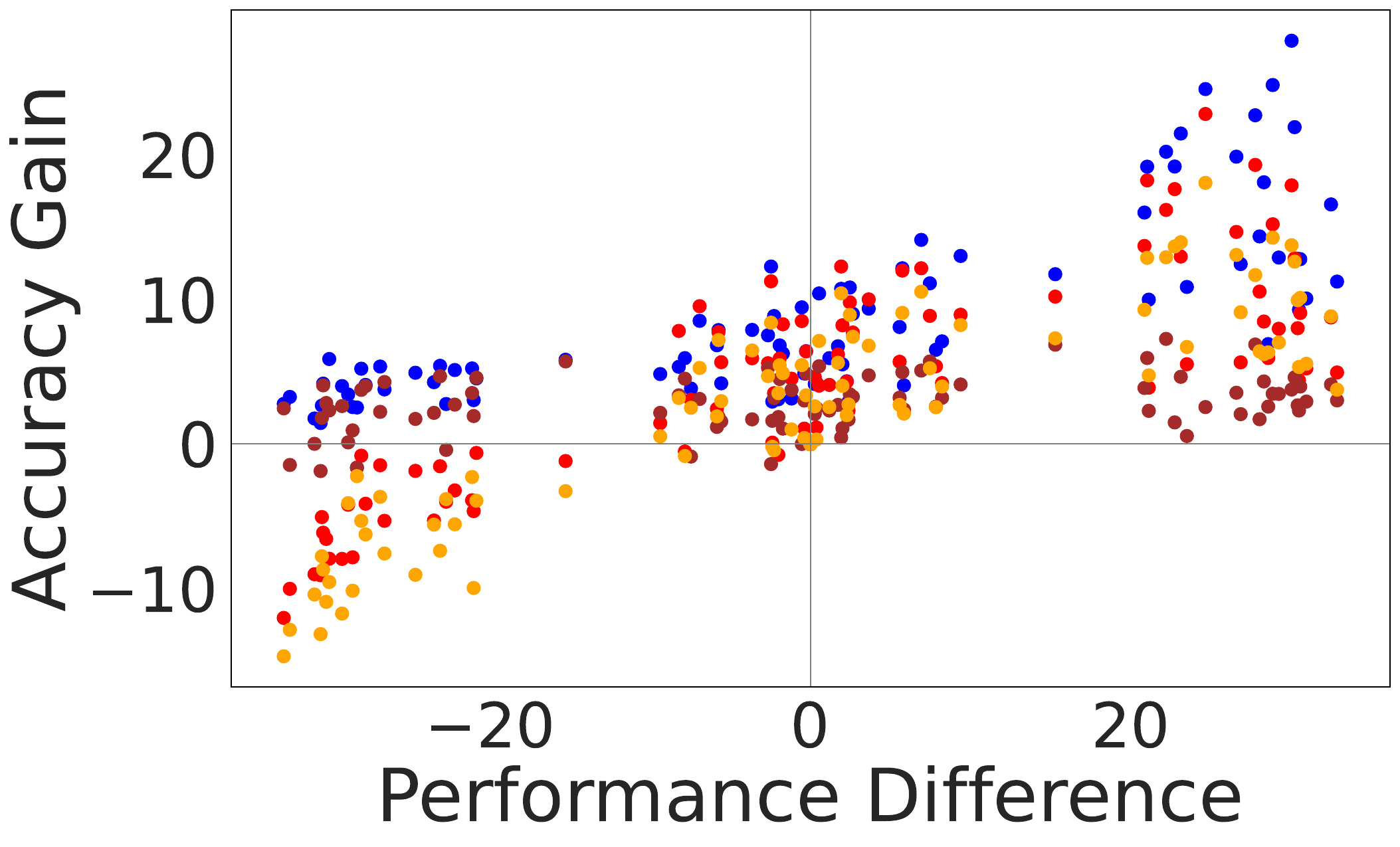} \end{adjustbox}\\
    \multicolumn{4}{c}{\textbf{\small{(e) Label Skew Dirichlet Distribution}}} 

  \end{tabularx}
    \caption{Optimization of hyperparameters \(  T \)  and \(\alpha\) in KD across various data partitioning strategies for the CIFAR-10 dataset. The first column illustrates the best \( T \) values derived from grid search, related to performance differences between student and teacher models. The second column shows the best \(\alpha\) values. The third column presents cumulative accuracy gains in percentage points from different KD methods. The final column details the accuracy gains for each participant pair using various KD techniques.}
  \label{fig:tunedKD}
\end{figure*}

\begin{figure*}[h!]
  \centering
  \begin{tabularx}{0.85\textwidth}{XXXX}
    \begin{adjustbox}{height=0.98in, keepaspectratio} 
    \includegraphics[width=\linewidth]{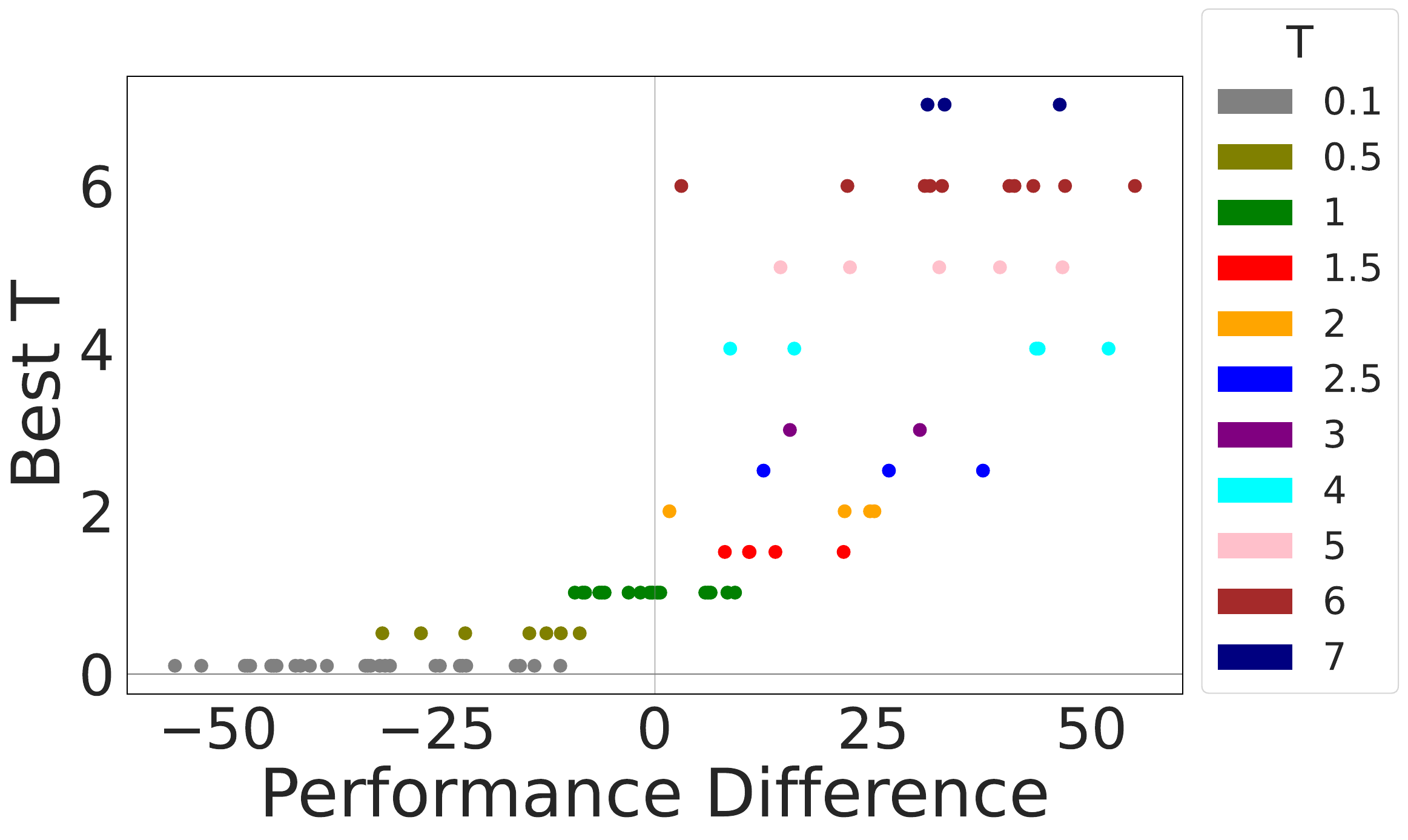} \end{adjustbox}&
    \begin{adjustbox}{height=0.9in, keepaspectratio} 
    \includegraphics[width=\linewidth]{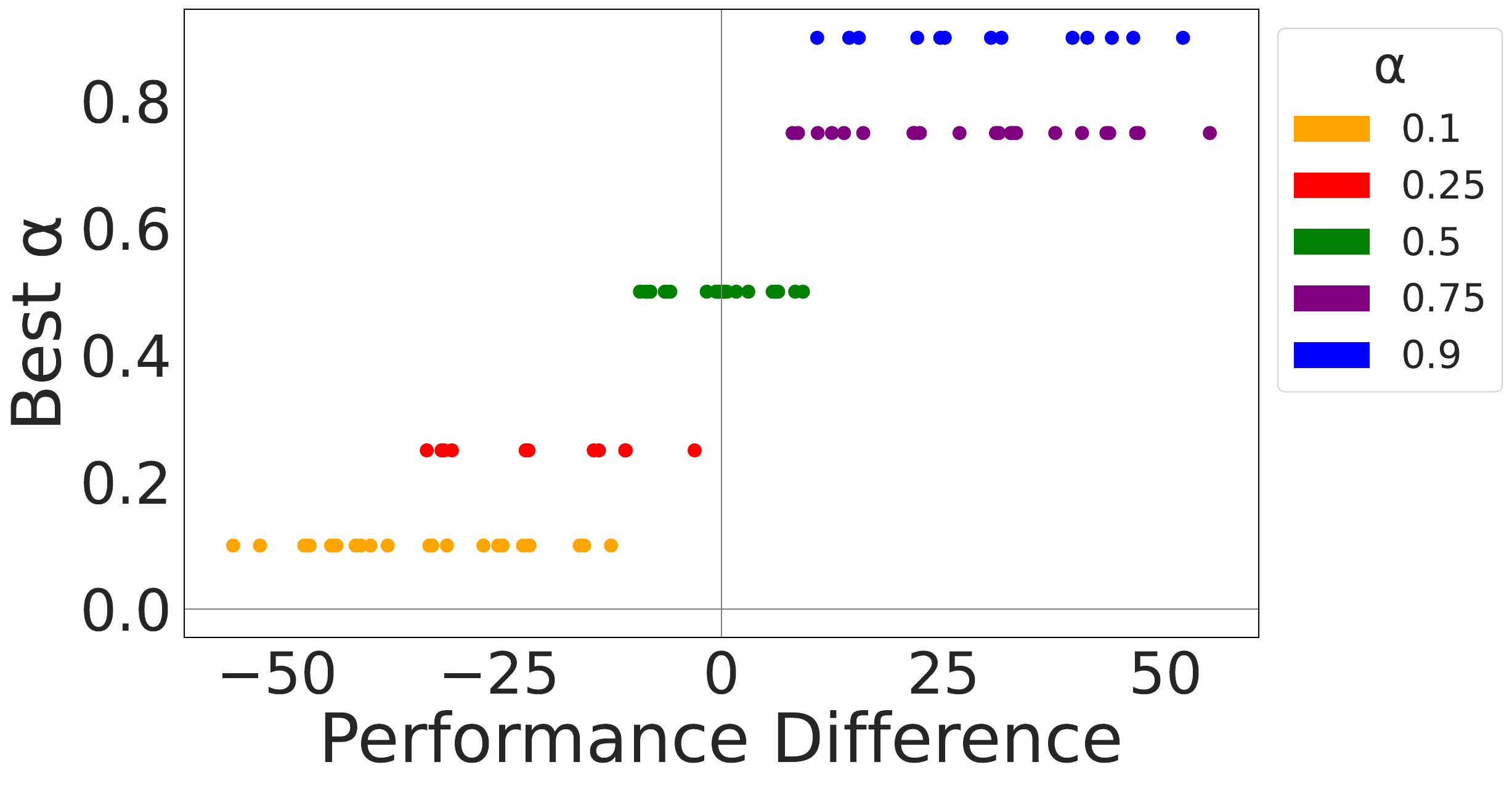} \end{adjustbox}&
    \begin{adjustbox}{height=0.9in, keepaspectratio}
    \hspace{7mm} 
    \includegraphics[width=\linewidth]{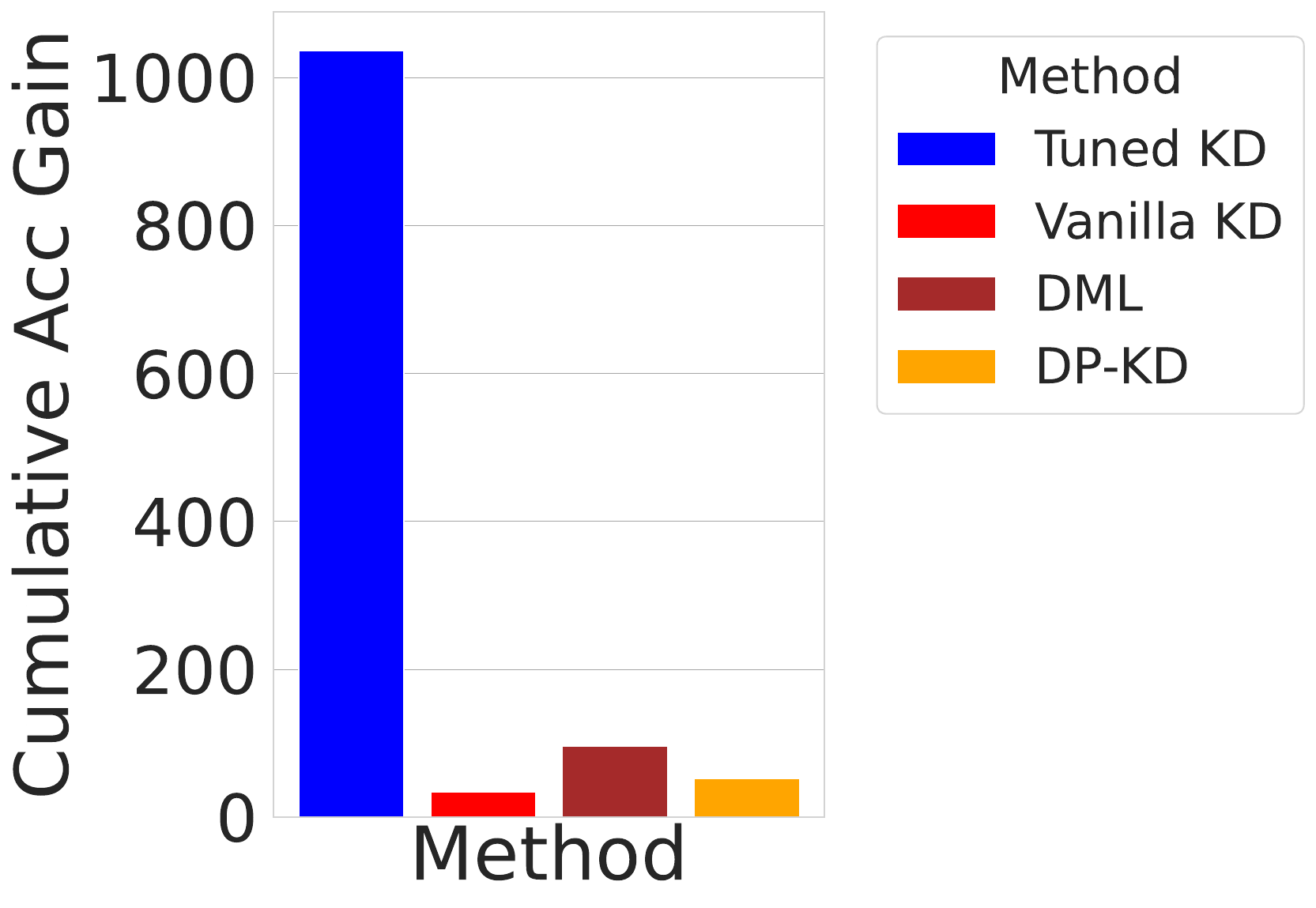} \end{adjustbox}&
    \begin{adjustbox}{height=0.9in, keepaspectratio}
    \includegraphics[width=\linewidth]{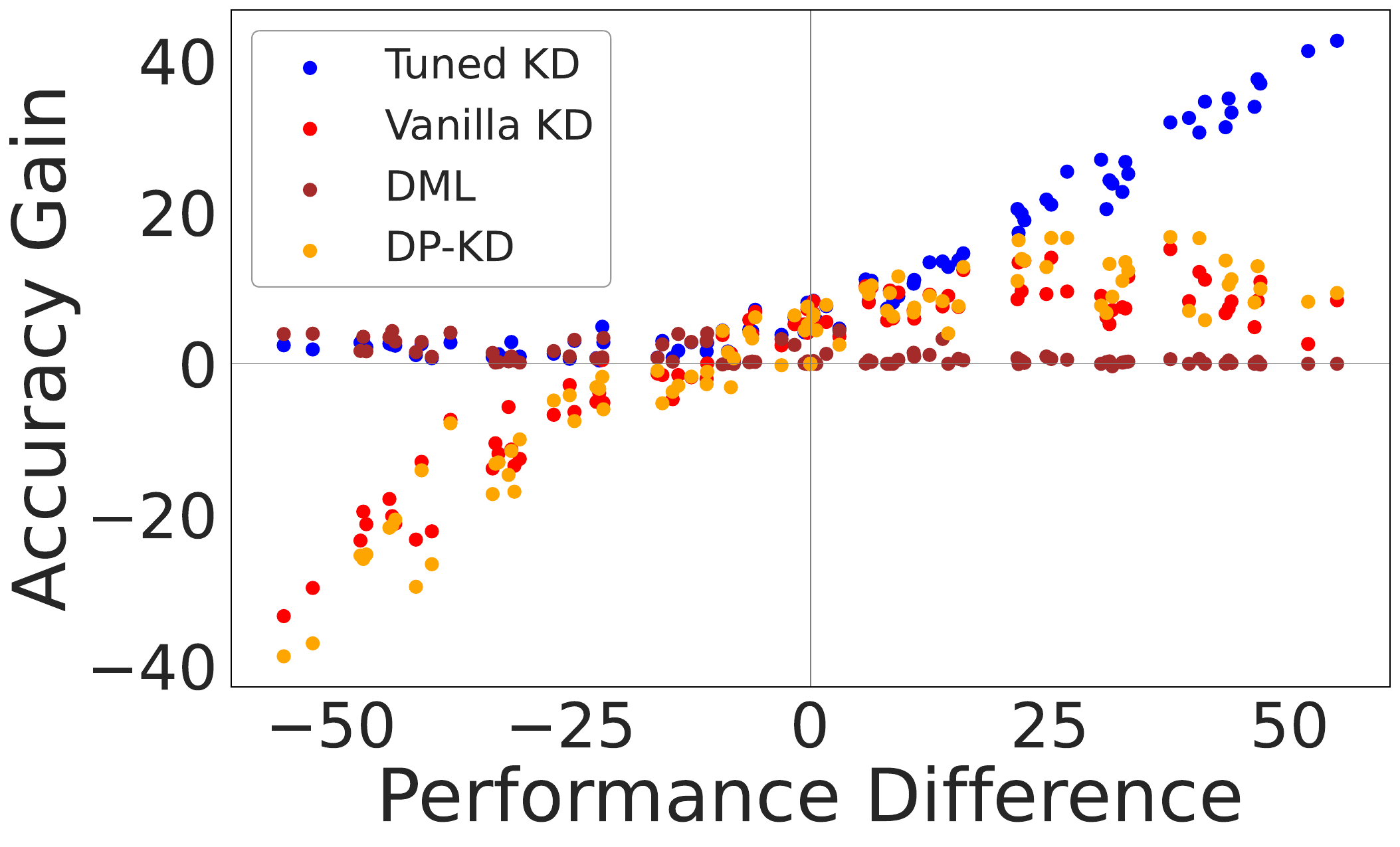} \end{adjustbox}\\

  \end{tabularx}
    \caption{Optimization of hyperparameters \(  T \)  and \(\alpha\) in KD for CINIC-10 dataset using Label Skew Random Chunks Distribution.}

  \label{fig:tunedKD_cinic10}
\end{figure*}

The optimization of hyperparameters is crucial in enhancing the efficiency of KD, a notion supported by recent research \cite{alballa2023first}. This subsection focuses on the efficacy of the Tuned KD approach, specifically targeting the optimization of the weight \( \alpha \) and then the temperature \( T \). This optimization process is conducted via a grid search over the parameter space outlined in \cref{tab:space}. For this analysis, and in the subsequent subsections unless otherwise noted, we use student data as the transfer set.

\cref{fig:tunedKD} illustrates the main outcomes of our hyperparameter optimization endeavors. It highlights the impact of varying \( T \) and \( \alpha \) values on KD performance across different data partitioning strategies and identifies appropriate settings for these parameters.

\begin{findingbox}
\textbf{Finding (6):} Tuned KD outperforms all the other KD approaches across all data partitioning strategies.
\end{findingbox}
\vspace{-2mm}
As demonstrated in the last column of \cref{fig:tunedKD}, Tuned KD consistently outperforms Vanilla KD, DML, and DP-KD in all data distribution scenarios. It is particularly noteworthy that Tuned KD achieves positive knowledge transfers even with weak teachers, markedly differing from the outcomes seen with Vanilla KD and DP-KD. Tuned KD enhances accuracy gains in situations involving strong teachers, surpassing all other KD approaches. Furthermore, the performance analysis summarized in the third column of \cref{fig:tunedKD} depicts the cumulative accuracy gain achieved by each KD method across different data distributions, reinforcing the standout performance of Tuned KD. 
This finding underscores the importance of optimally tuning the \( T \) and \( \alpha \) parameters.  It also opens up new avenues for future research, particularly in developing methods for efficient and automated tuning of these parameters tailored to the specific setting and learning objectives.

\begin{figure*}[htbp!]
\centering
\includegraphics[width=0.8\linewidth]{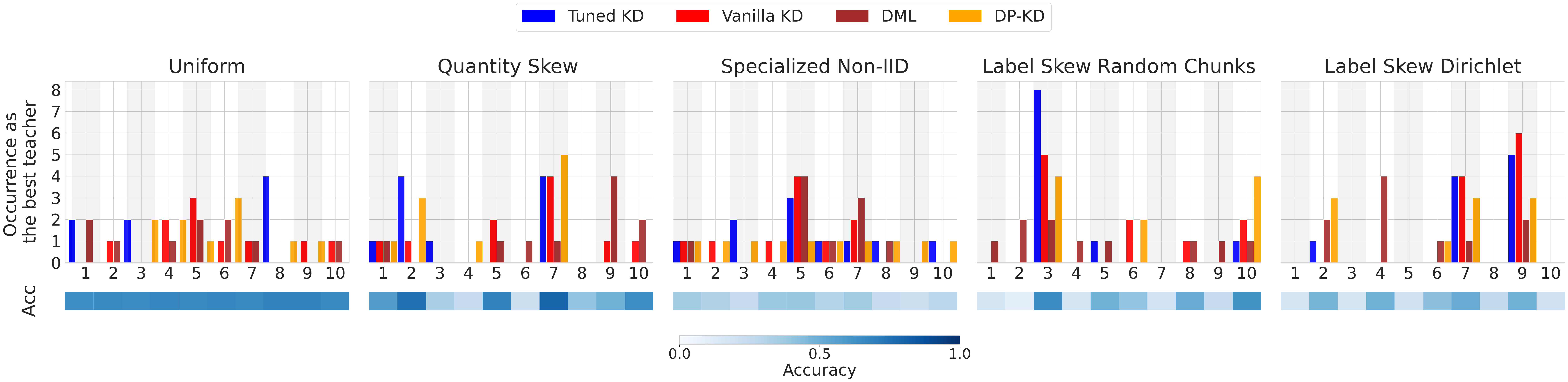}
\caption{Comparative analysis of the best teachers across diverse KD scenarios. The bar charts in the first row depict the frequency of each participant being selected as the best teacher within four different KD approaches across five data distribution strategies. The heatmaps in the second row illustrate the corresponding participant accuracies, providing insight into the relationship between a participant's accuracy and their likelihood of being an optimal teacher. The x-axis of the first row represents the participant' IDs, which align directly above their respective accuracy heatmap.}
\label{fig:best_teachers}
\end{figure*}

\begin{findingbox}
\textbf{Finding (7):} Tuning \( T \) and \( \alpha \) parameters becomes particularly important when the student and the teacher have diverse performances, while it becomes less significant when they have comparable performances. 
\end{findingbox}

The last column of \cref{fig:tunedKD} illustrates this phenomenon clearly. In scenarios with a significant performance disparity between the teacher and the student, Tuned KD demonstrates a notable improvement over Vanilla KD. Conversely, in cases where the teacher and student exhibit similar performance levels, the benefits of tuning \( T \) and \( \alpha \) diminish, resulting in Tuned KD achieving performance comparable to Vanilla KD's.

\begin{findingbox}
\textbf{Finding (8):}  Insights about the best \( T \) and \( \alpha \) values.
\end{findingbox}

\cref{fig:tunedKD}, particularly columns 1 and 2, illustrates the optimal \( T \) and \( \alpha \) values determined through our grid search across diverse data distributions. While identifying a consistent pattern for the best \( T \) and \( \alpha \) values in the Uniform Data Distribution is challenging, clear trends emerge in the Quantity Skew Distribution, the Label Skew Random Chunks Distribution, and the Label Skew Dirichlet Distribution. In these scenarios, a smaller \( T \) value tends to be more effective with a weaker teacher, while a larger \( T \) value is advantageous for a stronger teacher. Correspondingly, the ideal \( \alpha \) value, indicating the weight assigned to the teacher's output, should align with the teacher's relative strength compared to the student. Thus, a significantly stronger teacher warrants a higher \( \alpha \) value (above 0.5), while a weaker teacher should receive a lower weight. Moreover, the Specialized Non-IID Distribution favors lower values of \( T \) and \( \alpha \), leading to more favorable outcomes across all teacher-student pairings. 

\cref{fig:tunedKD_cinic10} shows the results from the experiments using an alternative practical CINIC-10 dataset \cite{darlow2018cinic10} to validate and reinforce our findings from the CIFAR-10 dataset. In these experiments, we opt for the challenging `Label Skew Random Chunks' data distribution approach. The results indicate the consistency of our observations drawn from CIFAR-10 results and suggest the potential generalizability of our conclusions about the efficacy of tuning across different benchmarks, especially in diverse data environments. 

\subsection{Best teachers across KD methods and data distributions}

\cref{fig:best_teachers} juxtaposes the frequency of participants emerging as the best teachers against their respective performance within varied data distribution strategies, utilizing the four KD approaches. The best teacher is defined as one who provides the greatest accuracy gain for a given student participant. This analytical approach aims to discern the potential correlation between a participant's accuracy and their propensity to be an effective teacher.

\begin{findingbox}
\textbf{Finding (9):} Participant's accuracy is not the sole determinant of its efficacy as a teacher in KD.
\end{findingbox}

The findings indicate that participant accuracy, though pivotal, is not the only factor influencing the selection of the best teacher across diverse KD techniques and data distributions. This highlights the complexity of the KD teacher selection process, which considers a broader spectrum of factors.

Under the uniform data distribution, the emergence of the best teachers appears to be evenly spread across participants for most KD approaches. 
In contrast, in scenarios with significant data skewness, particularly in quantity and label skew distributions, KD methods such as Tuned KD, Vanilla KD, and DP-KD tend to prefer highly accurate teachers, but also possibly incorporate other factors in their selection.

Interestingly, the DML approach's teacher selection exhibits a weaker correlation with accuracy, implying that it may value additional attributes such as the diversity or specificity of participant knowledge. This intricate selection dynamic indicates an opportunity to enhance KD algorithms by integrating a more complex array of criteria beyond accuracy. Subsequent studies should aim to decode these implicit factors, potentially leading to more refined and efficacious KD paradigms.

\subsection{Model Learning and Forgetting Dynamics}
 
We examine the relation between KD methods and the model's learning and forgetting under different data distribution strategies, consolidating our earlier observations and unveiling new insights into the adaptability and robustness of these methods.

\begin{figure*}[htbp]
  \centering
  \begin{subfigure}[b]{0.9\textwidth}
    \centering
    \includegraphics[width=\linewidth]{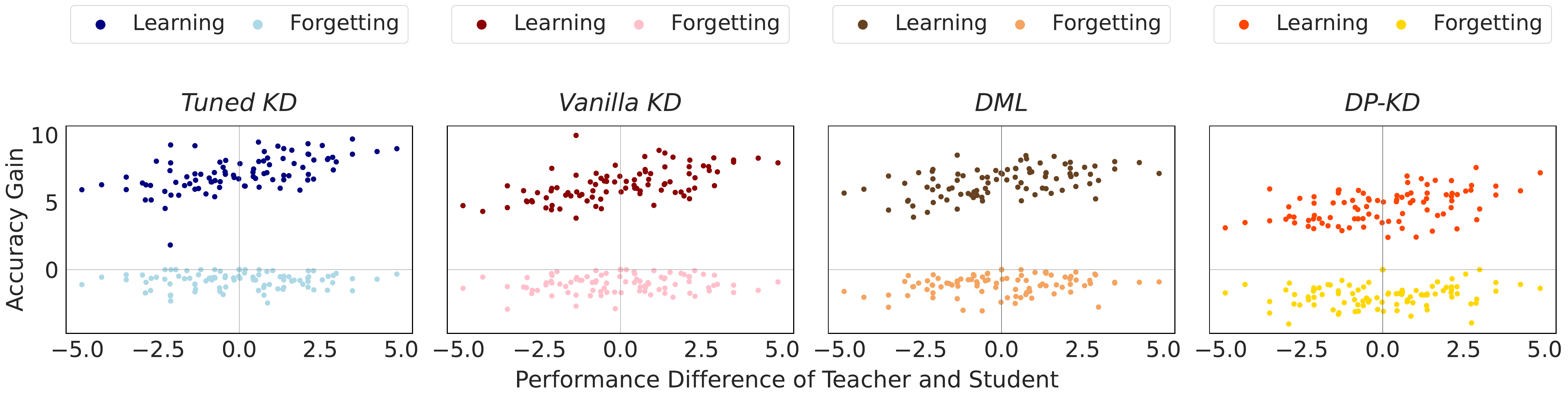}
    \caption{\small{Uniform Data Distribution}}
  \end{subfigure}
  \vspace{4mm}
  
  \begin{subfigure}[b]{0.9\textwidth}
    \centering
    \includegraphics[width=\linewidth]{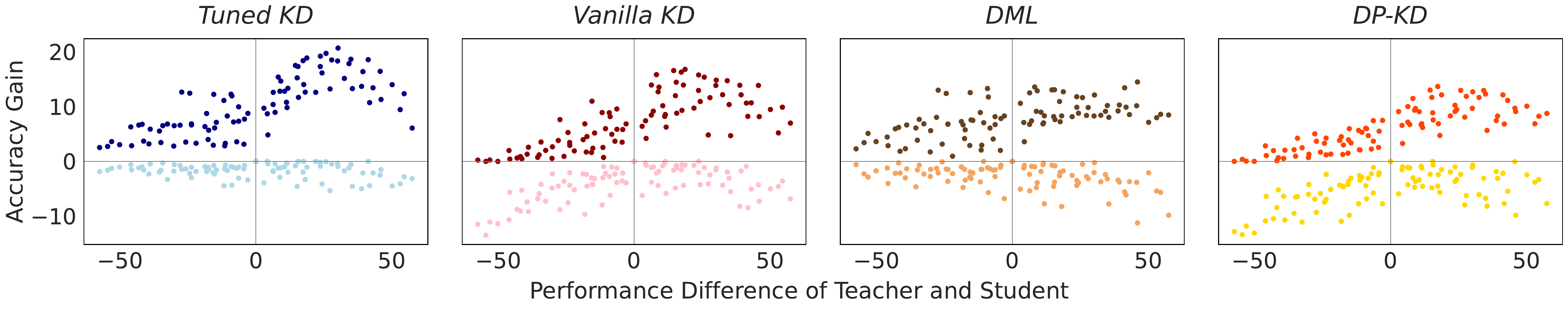}
    \caption{\small{Quantity Skew Distribution}}
  \end{subfigure}
  \vspace{4mm}
  
  \begin{subfigure}[b]{0.9\textwidth}
    \centering
    \includegraphics[width=\linewidth]{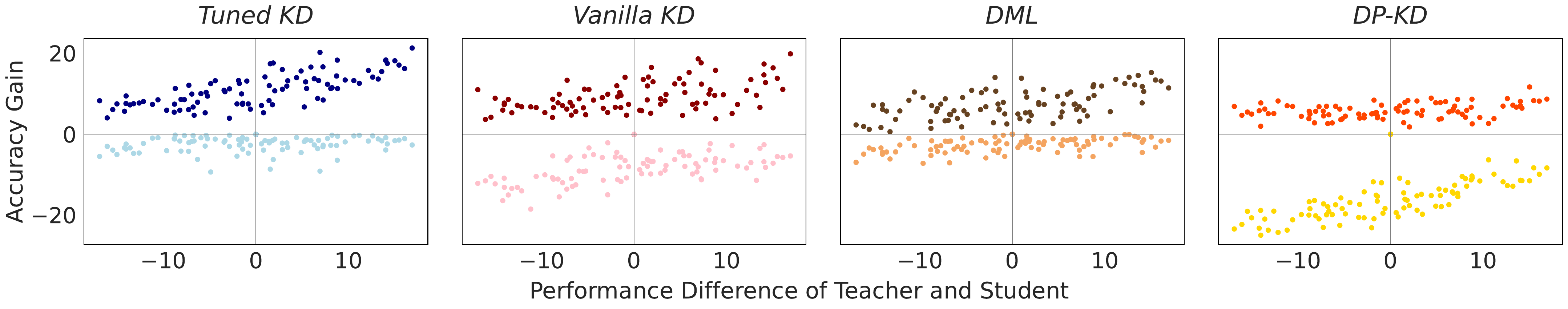}
    \caption{\small{Specialized Non-IID Distribution}}
  \end{subfigure}
  \vspace{4mm}
  
  \begin{subfigure}[b]{0.9\textwidth}
    \centering
    \includegraphics[width=\linewidth]{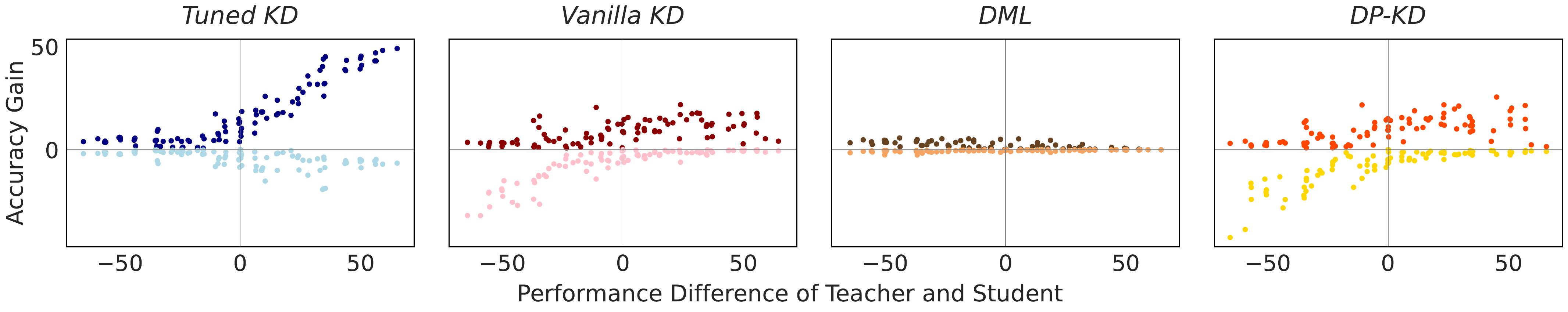}
    \caption{\small{Label Skew Random Chunks Distribution}}
  \end{subfigure}
  \vspace{4mm}

   \begin{subfigure}[b]{0.9\textwidth}
    \centering
    \includegraphics[width=\linewidth]{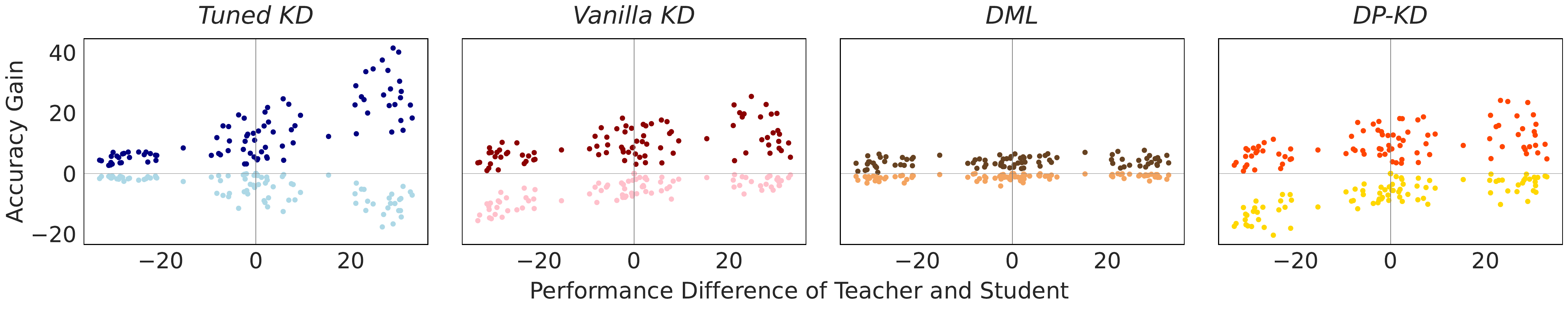}
    \caption{\small{Label Skew Dirichlet Distribution}}
  \end{subfigure}
  \vspace{3mm}

  \caption{Comparative analysis of learning vs. forgetting in KD across diverse data distributions.}
  \label{fig:learningVSforgetting}
\end{figure*}

\cref{fig:learningVSforgetting} investigates the learning-forgetting (cf. \cref{sec:methodology}) trade-off in KD methods. Tuned KD and DML show less forgetting with weaker teachers, outperforming Vanilla KD and DP-KD. Uniform data distributions reveal similar performances across methods, with a minor uptick in forgetting for DP-KD. Tuned KD emerges as the most effective, maximizing learning in all scenarios and reducing forgetting in Quantity Skew and Specialized Non-IID distributions. In contrast, it shows increased forgetting within Label Skew distributions. DML shows minimal forgetting but is constrained in learning gains, suggesting a conservative knowledge consolidation preference.
This analysis underscores the need to tailor the KD approach selection to each method's specific strengths and weaknesses and align it with data characteristics and learning goals.

\subsection{Influence of Data Distribution and Design Choices for KD}

\begin{figure}[htbp!]
\centering
\includegraphics[width=1\linewidth]{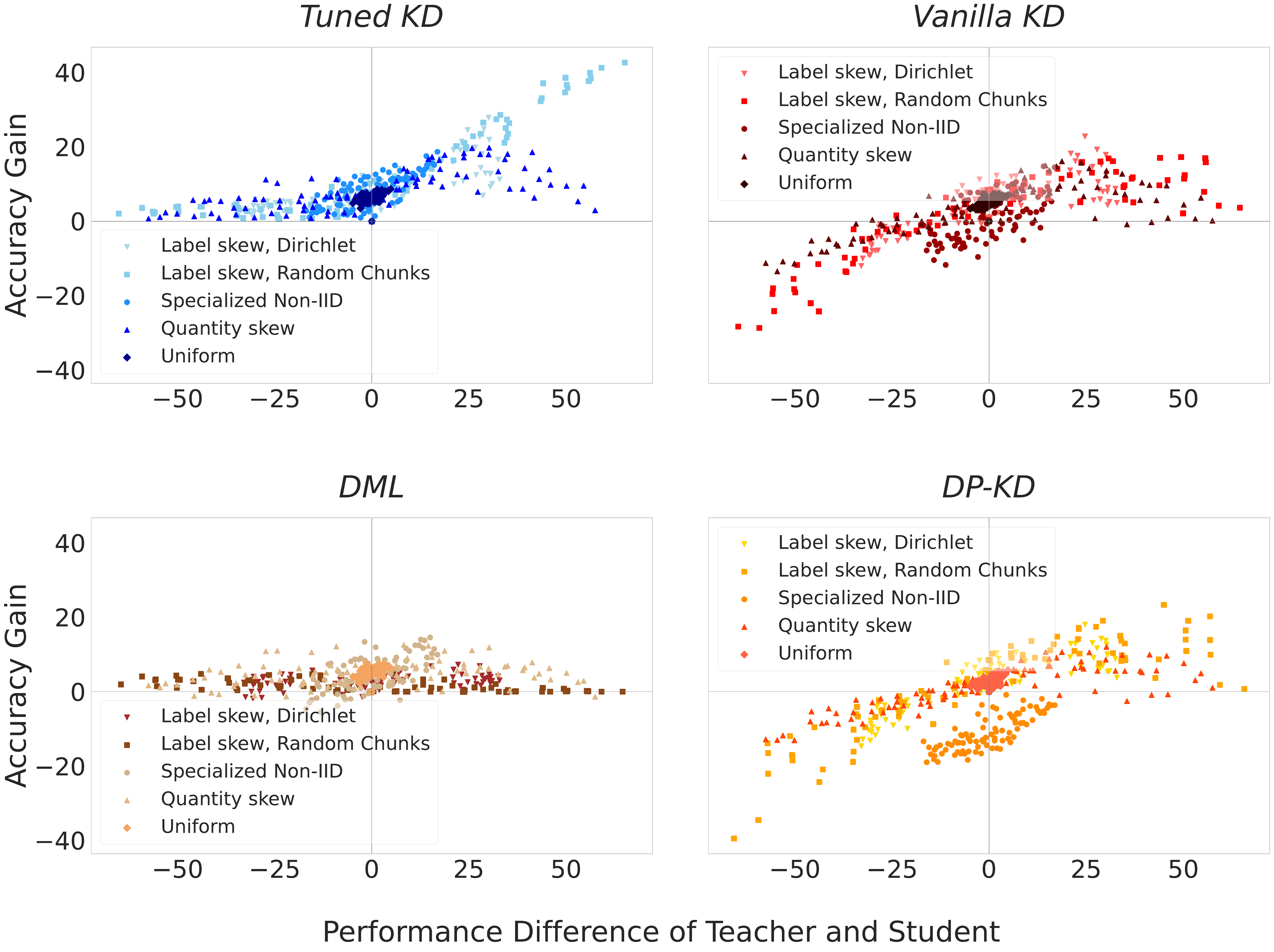}
\caption{Knowledge distillation techniques against varied data distribution characteristics.}
\label{fig:KDsvsDataDist}
\end{figure}

We re-examine the performance of various KD methods, including Tuned KD, under a different light and unveil new insights into the adaptability and robustness of these methods. 
\cref{fig:KDsvsDataDist} juxtaposes the efficacy of KD methods against varied data distributions. This comparative visualization allows us to discern patterns and outliers in performance, highlighting the consistent strength of Tuned KD across multiple scenarios. By contrasting these methods side-by-side, we better understand how each method responds to diverse data conditions.

\begin{figure}[htbp!]
  \centering
  \begin{mdframed}[backgroundcolor=white, linecolor=gray, innertopmargin=7pt, innerbottommargin=7pt, innerleftmargin=7pt, innerrightmargin=7pt, roundcorner=2pt]
    \includegraphics[width=1\linewidth]{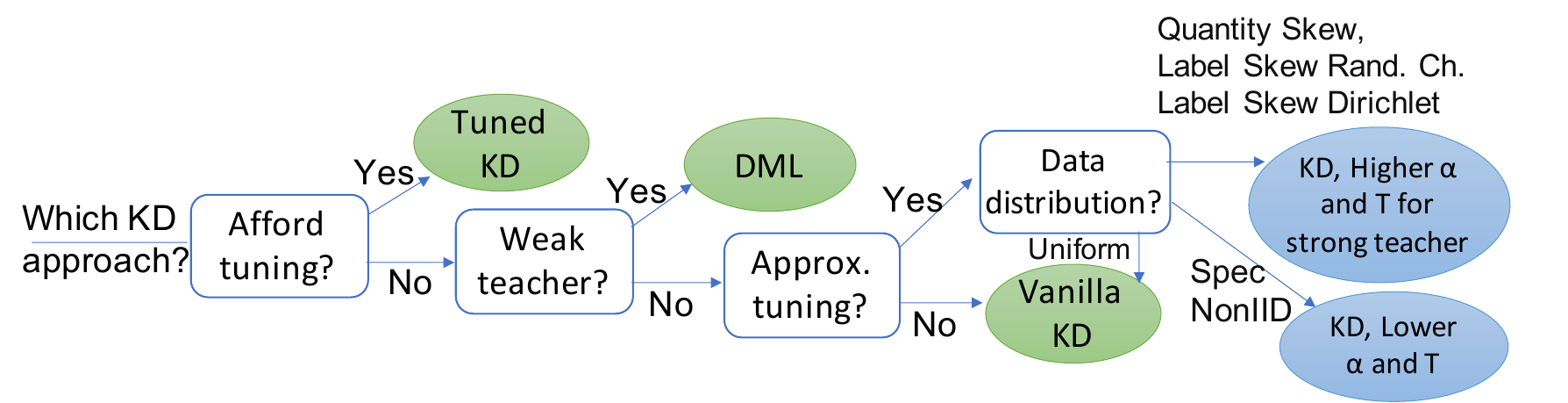}
  \end{mdframed}
  \caption{Decision tree to determine the suitable KD approach.}
  \label{fig:Dec tree}
\end{figure}

Finally, based on the previous findings, we create the decision tree shown in \cref{fig:Dec tree} as a strategic guide for selecting suitable KD methods based on specific use cases and data environments, offering a practical approach for applying KD methods in various contexts.

\begin{table*}[ht]
\centering
\tiny
\caption{Results of combining multiple models across various data distributions and transfer set options. The table presents accuracy in percentage points post-distillation. }
\label{tab:kd_comparison_noES}
\footnotesize 
\renewcommand{\arraystretch}{1} 
\setlength{\tabcolsep}{3pt} 
\begin{tabularx}{\textwidth}{@{}X*{10}{c}@{}} 
\toprule
\multirow{2}{*}{Data Distribution} & \multicolumn{6}{c}{Public Unlabeled Transfer Set} & \multicolumn{4}{c}{Student Data as Transfer Set} \\
\cmidrule(lr){2-7} \cmidrule(l){8-11}
 & \multicolumn{2}{c}{Worst-Student} & \multicolumn{2}{c}{Best-Student} & \multicolumn{2}{c}{Untrained-Model} & \multicolumn{2}{c}{Worst-Student} & \multicolumn{2}{c}{Best-Student} \\
 & Weighted & Unweighted & Weighted & Unweighted & Weighted & Unweighted & Weighted & Unweighted & Weighted & Unweighted \\
\midrule
Uniform & 71.64 & 72.36 & 72.35 & \textbf{73.43} & 69.55 & 70.59 & 71.34 & 72.24 & 72.68 & \textbf{73.97} \\
Quantity Skew & 68.18 & 64.89 & \textbf{77.90} & 73.86 & 67.31 & 64.66 & 29.77 & 30.01 & \textbf{78.34} & 76.33 \\
Specialized Non-IID & 51.57 & 54.72 & 53.07 & \textbf{55.11} & 52.55 & 52.98 & 43.70 & 50.72 & 52.39 & \textbf{53.60} \\
Label Skew Random Chunks & 65.44 & 66.73 & \textbf{73.33} & 71.68 & 66.35  & 66.22 & 19.22 & 20.67 & \textbf{75.71} & 75.41 \\
Label Skew Dirichlet & 61.67 & 54.03 & \textbf{67.63} & 57.82 & 63.77 & 52.41 & 28.61 & 27.35 & \textbf{66.92} & 63.00 \\
\bottomrule
\end{tabularx}
\end{table*}

\subsection{KD from Multiple Pre-trained Models}
\label{FL}
This subsection evaluates the application of KD in the context of multiple pre-trained models. Our investigation primarily revolves around Vanilla KD, examining its effectiveness in consolidating knowledge from various models, each with distinct training experiences and data exposures. 

This analysis aligns with the broader goal of harnessing the collective intelligence of diverse models in centralized collaborative learning and FL systems.

Additionally, we examine how KD can be strategically employed to pre-consolidate knowledge across models before they participate in the FL process. By doing so, we aim to streamline the FL workflow, potentially reducing the number of communication rounds needed and accelerating the convergence toward optimal model performance.

For FL experiments, we follow the optimized Federated Averaging (FedAvg) parameters from \cite{caldarola2022improving}, chosen for their relevance to our CIFAR-10 dataset and ResNet-18 model setup. We use an SGD solver with a learning rate of 0.01, weight decay of $4 \times 10^{-4}$, no momentum, and local epochs set to 2.
We implement a standard cross-silo FL setting with full participation (participation rate of 1) across 100 communication rounds.

\subsubsection{\textbf{Combining multiple models}}
We here delve into methodologies for the effective amalgamation of knowledge across multiple pre-trained models via KD. Our focus includes critical aspects such as the selection of an appropriate starting student model and the assessment of various teacher weighting strategies across different transfer set scenarios.

\smartparagraph{Where to start?}
Selecting the starting student model is crucial in knowledge distillation. We explore three scenarios to determine the most effective approach for initiating the student model for the amalgamation process: 1) the worst-performing model among the group; 2) the best-performing model; or 3) an untrained student model, which begins from a completely randomized, untrained state.

\smartparagraph{Transfer set.}
Our approach incorporates two transfer set options: 1) a public unlabeled dataset with 5,000 samples; and 2) the student's dataset. The models undergo 30 epochs of training over these transfer sets.

\begin{figure}[!t]
    \centering
    \includegraphics[width=0.26\textwidth]{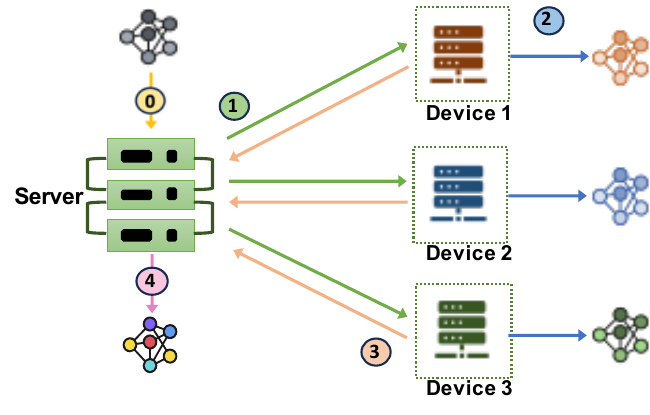}
    \caption{FL Process with Pre-Consolidated Model Initialization. The server initializes the global model using the pre-consolidated model (Step 0), which incorporates knowledge from multiple teacher models. Step 0 occurs only once at the beginning. The subsequent steps (1-4) follow the standard FL process: the global model is sent to clients (Step 1), updated locally by each client (Step 2), sent back to the server (Step 3), and aggregated to form a new global model (Step 4). This process is repeated for a predefined number of rounds or until a certain performance criterion is met.}
    \label{fig:FL_process}
\end{figure}

 \begin{figure*}[htbp!]
\centering
\includegraphics[width=0.95\linewidth]{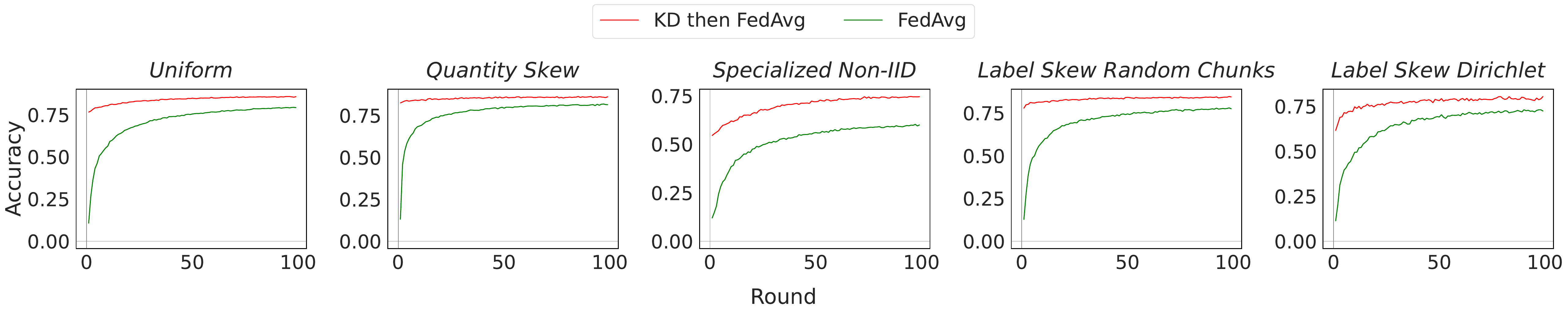}
\vspace{-1mm}
\caption{Comparative performance of traditional FedAvg vs. KD Pre-consolidated FedAvg across various data partitioning strategies.}
\label{fig:FedAvg_VS_combined_then_FedAvg}
\end{figure*}

\cref{tab:kd_comparison_noES} summarises our experimental outcomes, offering insights into the efficacy of different KD strategies under varying conditions. The `Weighted' and the `Unweighted' columns in \cref{tab:kd_comparison_noES} represent the adaptive weighting and the equal weighting strategies, respectively (cf. \cref{eqn:distillation-multi-kl}).

\begin{findingbox}
\textbf{Finding (10):} Selecting the highest-performing model as the student in the consolidation process consistently yields the most favorable outcomes.
\end{findingbox}
The data show a consistent trend: initiating the distillation process with the best-performing student model leads to better results. This can be due to its own developed knowledge. 

\begin{findingbox}
\textbf{Finding (11):} The adaptive weighting approach shows varied efficacy based on data distribution, with marked gains in Label Skew Dirichlet and Quantity Skew scenarios.
\end{findingbox}
The adaptive weighting approach delineated in \cref{eqn:dw_multi} and \cref{eqn:distillation-multi-kl} manifests considerable enhancements in specific data distribution contexts, notably the Label Skew Dirichlet and Quantity Skew scenarios. Conversely, the equal weighting strategy across all teachers appears more beneficial in the Specialized Non-IID and the Uniform data distributions.
This variability underscores the need to tailor weighting strategies based on specific data contexts in KD applications. 

Moreover, the choice between utilizing the best student data or a public unlabeled dataset as the transfer set did not show significant differences in performance, suggesting flexibility in transfer set selection based on availability. 

However, comparing using the best and worst student data as the transfer set shows a marked difference. Opting for the best student data typically leads to significantly better results. This can be attributed to the broader and possibly richer data representation in the best student's dataset, in contrast to the more limited scope of the worst student's data. This finding highlights the importance of transfer set size and quality, corroborating the initial observations we made in \cref{sec:eval-kd} regarding the impact of transfer set on KD performance.

Given these observations, we proceed with further experiments using the student data as the transfer set and employ the adaptive weighting mechanism for knowledge distillation.

\subsubsection{\textbf{Accelerated FL with KD pre-consolidation}}
\label{FL_accel}

We introduce a novel method of using KD for model pre-consolidation to enhance centralized collaborative training, particularly in cross-silo FL settings. Our approach deviates from traditional FL practices that typically begin with randomly initialized models. Instead, it involves a preliminary phase where multiple models from different silos are first consolidated using KD, as depicted in \cref{fig:pre_consolidation}. This pre-consolidation process entails distilling knowledge from various pre-trained models into a single, unified model. This pre-consolidated model serves as the starting point for federated learning (\cref{fig:FL_process}), aiming to reduce the heterogeneity typically encountered at the onset of traditional FL.

We compare traditional FL using FedAvg against our proposed approach, which utilizes the pre-consolidated model as the initial model for FL. Our analysis spans various data distribution scenarios to gauge the impact of KD pre-consolidation on reducing the number of communication rounds necessary to achieve target accuracy in FL. \cref{fig:FedAvg_VS_combined_then_FedAvg} presents the results of this comparative analysis.

\begin{findingbox}
\textbf{Finding (12):} Employing KD for model pre-consolidation as an initial step in FL can significantly accelerate the learning process.
\end{findingbox}

When used alone, KD consolidation might not always produce satisfactory results in complex environments like Specialized Non-IID and Label Skew Dirichlet distributions. However, incorporating it as a preliminary step in FL significantly speeds up the learning process. This demonstrates the potential of KD pre-consolidation as a powerful tool for improving the efficiency of FL systems, particularly in accelerating the convergence toward target accuracy levels.

\section{Conclusion}
\label{conclusion}

This paper presents a detailed examination of knowledge distillation within collaborative learning contexts, particularly focusing on scenarios that employ pre-trained models. Our in-depth exploration provided crucial insights into optimizing KD techniques for enhancing collaborative learning effectiveness.

We have systematically compared various KD methodologies, such as Vanilla KD, Tuned KD, DML, and DP-KD, across multiple data distribution strategies, highlighting their unique strengths and applications. Our findings demonstrate that Tuned KD consistently outperforms other methods in various scenarios, particularly with optimized parameters.

Our research reveals the potential of KD to accelerate FL in the context of centralized collaborative learning systems. By pre-consolidating models using KD, our experiments indicate a substantial reduction in the number of communication rounds required for FL, thus enhancing its efficiency. Overall, our research enriches the understanding of KD within collaborative learning systems. 

\bibliographystyle{IEEEtran} 
\bibliography{ref}


\end{document}